\pdfoutput=1
\PassOptionsToPackage{usenames, dvipsnames}{xcolor}
\documentclass[acmsmall]{acmart} 

\usepackage{multirow}
\usepackage{subcaption}
\usepackage{comment}
\usepackage{float}
\usepackage[normalem]{ulem}
\useunder{\uline}{\ul}{}
\usepackage[usenames, dvipsnames]{xcolor}

\AtBeginDocument{%
  \providecommand\BibTeX{{%
    \normalfont B\kern-0.5em{\scshape i\kern-0.25em b}\kern-0.8em\TeX}}}

\setcopyright{acmcopyright}  
\copyrightyear{2022}
\acmYear{2022}
\acmDOI{10.1145/1122445.1122456}

\acmJournal{JACM}  
\acmVolume{37}
\acmNumber{4}
\acmArticle{111}
\acmMonth{8}
\begin{document}

%%
%% The "title" command has an optional parameter,
%% allowing the author to define a "short title" to be used in page headers.
\title[AABL]{Authorship Attribution in Bangla Literature (AABL) via Transfer Learning using ULMFiT}

%%
%% The "author" command and its associated commands are used to define
%% the authors and their affiliations.
%% Of note is the shared affiliation of the first two authors, and the
%% "authornote" and "authornotemark" commands
%% used to denote shared contribution to the research.
%=====================================================================================
\author{Aisha Khatun}
% \authornote{Both authors contributed equally to this research.}
\email{aysha.kamal7@gmail.com}
\orcid{0000−0001−6559−4493}
% \author{G.K.M. Tobin}
% \authornotemark[1]
% \email{webmaster@marysville-ohio.com}
\author{Anisur Rahman}
\email{emailforanis@gmail.com}
\orcid{0000−0002−4616−4559}
\author{Md Saiful Islam}
\email{saiful-cse@sust.edu}
\orcid{0000−0001−9236−380X}
\author{Hemayet Ahmed Chowdhury}
\email{hemayetchoudhury@gmail.com}
\orcid{0000−0002−5582−669X}
\author{Ayesha Tasnim}
\email{tasnim-cse@sust.edu}
\orcid{0000−0002−7143−3255}
\affiliation{%
  \institution{Department of Computer Science and Engineering, Shahjalal University of Science and Technology}
  \streetaddress{Kumargaon}
  \city{Sylhet}
  \state{Bangladesh}
  \postcode{3114}
}

% \author{Lars Th{\o}rv{\"a}ld}
% \affiliation{%
%   \institution{The Th{\o}rv{\"a}ld Group}
%   \streetaddress{1 Th{\o}rv{\"a}ld Circle}
%   \city{Hekla}
%   \country{Iceland}}
% \email{larst@affiliation.org}

% \author{Valerie B\'eranger}
% \affiliation{%
%   \institution{Inria Paris-Rocquencourt}
%   \city{Rocquencourt}
%   \country{France}
% }

% \author{Aparna Patel}
% \affiliation{%
%  \institution{Rajiv Gandhi University}
%  \streetaddress{Rono-Hills}
%  \city{Doimukh}
%  \state{Arunachal Pradesh}
%  \country{India}}

% \author{Huifen Chan}
% \affiliation{%
%   \institution{Tsinghua University}
%   \streetaddress{30 Shuangqing Rd}
%   \city{Haidian Qu}
%   \state{Beijing Shi}
%   \country{China}}

% \author{Charles Palmer}
% \affiliation{%
%   \institution{Palmer Research Laboratories}
%   \streetaddress{8600 Datapoint Drive}
%   \city{San Antonio}
%   \state{Texas}
%   \postcode{78229}}
% \email{cpalmer@prl.com}

% \author{John Smith}
% \affiliation{\institution{The Th{\o}rv{\"a}ld Group}}
% \email{jsmith@affiliation.org}

% \author{Julius P. Kumquat}
% \affiliation{\institution{The Kumquat Consortium}}
% \email{jpkumquat@consortium.net}
%======================================================================================

%%
%% By default, the full list of authors will be used in the page
%% headers. Often, this list is too long, and will overlap
%% other information printed in the page headers. This command allows
%% the author to define a more concise list
%% of authors' names for this purpose.
%======================================================================================
\renewcommand{\shortauthors}{Khatun, et al.}
%======================================================================================

%%
%% The abstract is a short summary of the work to be presented in the
%% article.
\begin{abstract}
Authorship Attribution is the task of creating an appropriate characterization of text that captures the authors' writing style to identify the original author of a given piece of text. With increased anonymity on the internet, this task has become increasingly crucial in various security and plagiarism detection fields. Despite significant advancements in other languages such as English, Spanish, and Chinese, Bangla lacks comprehensive research in this field due to its complex linguistic feature and sentence structure. Moreover, existing systems are not scalable when the number of author increases, and the performance drops for small number of samples per author. In this paper, we propose the use of Average-Stochastic Gradient Descent Weight-Dropped Long Short-Term Memory (AWD-LSTM) architecture and an effective transfer learning approach that addresses the problem of complex linguistic features extraction and scalability for authorship attribution in Bangla Literature (AABL). We analyze the effect of different tokenization, such as word, sub-word, and character level tokenization, and demonstrate the effectiveness of these tokenizations in the proposed model. Moreover, we introduce the publicly available Bangla Authorship Attribution Dataset of 16 authors (BAAD16) containing 17,966 sample texts and 13.4+ million words to solve the standard dataset scarcity problem and release six variations of pre-trained language models for use in any Bangla NLP downstream task. For evaluation, we used our developed BAAD16 dataset as well as other publicly available datasets. Empirically, our proposed model outperformed state-of-the-art models and achieved 99.8\% accuracy in the BAAD16 dataset. Furthermore, we showed that the proposed system scales much better even with an increasing number of authors, and performance remains steady despite few training samples.
\end{abstract}

%======================================================================================
%%
%% The code below is generated by the tool at http://dl.acm.org/ccs.cfm.
%% Please copy and paste the code instead of the example below.
%%



%======================================================================================
%%
%% Keywords. The author(s) should pick words that accurately describe
%% the work being presented. Separate the keywords with commas.

%%
%% This command processes the author and affiliation and title
%% information and builds the first part of the formatted document.
\maketitle

\section{Introduction}

Authorship attribution (AA) is a distinct type of classification task that deals with identifying the author of an anonymous piece of writing within a set of probable authors. Identifying the original author depends on capturing the elusive characteristics of an author's writing style from their digitized texts by analyzing stylometric and linguistic features. 
Despite extensive use of stylometry to identify the literary styles of authors, AA is challenging because of
(i) the complex structure of language 
(ii) similarity of the topic of discussion,
(iii) implicit writing styles of authors,
(iv) known forms of writing, such as novels or stories, often have similar structures.
Extracting useful information about an author's style is difficult, especially to build an end-to-end system to detect the author of a text.

Anonymity is widespread in recent times, primarily due to the widespread use of the internet; the use and misuse of anonymity have become an essential factor to consider. The plethora of unattributed digital footprints makes authorship attribution indispensable in various fields, and its applications are constantly growing. Application of authorship attribution covers sectors such as forensic investigation, plagiarism detection, computer security, criminal law, cyber-crime, literature, trading, etc. Intelligence agencies can use it to link intercepted messages to known enemies; original authors of harassing messages can be identified. In education, student submissions can be verified, and actual authors can be identified using authorship attribution.

Early approaches of AA followed statistical methods that suffered from text length limitation problems and failed to capture the text's semantic meaning. Most stylometric studies use some language items, such as lexical and syntactic components. Among them, character n-gram has been used widely and shown to provide insights into the writing style of the authors \cite{stamatatos2009survey}. In recent studies, machine learning approaches \cite{zhang2014authorship} are used to extract complex and essential features of the text to identify the original author's writing style. Some research extracted semantic information using pre-trained word, and character embeddings \cite{ourchar}. With increasing computing power, various deep learning models have also been successfully applied to the task, achieving impressive results \cite{jafariakinabad2019syntactic,ruder2016character}.

AA is not limited to any specific language and pertains to any written text. Bangla is an Indic language, and the 7th most spoken language globally, shared among the people of Bangladesh, the Indian states of West Bengal, Tripura, Assam, and the global Bangla diaspora communities \cite{wiki-bang}. Despite success in English and other languages \cite{jafariakinabad2019syntactic,ruder2016character, kreutz2018exploring, sundararajan2018represents}, Bangla lacks significant work in this area due to its high inflection "with more than 160 different inflected forms for verbs, 36 different forms for nouns, and 24 different forms for pronouns" \cite{bhattacharya2005inflectional}. Moreover, Bangla exhibits diglossia, Shadhu and Cholito forms: these are two different writing styles with slightly different vocabularies and syntax. Although the former is uncommon in modern writings, it constitutes a vast majority of classical literature in Bangla, and both forms share a large number of common roots. All these contribute to the complex written form of Bangla and its extended vocabulary. 

Only a few research work was done to conduct Authorship Attribution in Bangla Literature (AABL), but most of them lack in some important aspect. Most works apply traditional approaches to extract features like word length, sentence length, word use frequency, etc., along with n-grams \cite{das2015experimental, hossain2017stylometric, islam2018authorship, ahmad2020empirical}. Besides, embedding-based models have also been used to encode words \cite{bhai} and characters \cite{ourchar} to extract authorial information without any manual feature selection. Although a few promising studies have been done using deep learning, existing systems suffer due to the substantial dependence on the dataset. The limitations of the present systems are described below:

\begin{enumerate}
    \item Most AABL works use manual feature engineering, which is corpus dependent and labour-intensive. High-performing end-to-end systems are not widely available for AABL.
    \item Lack of resources and scarcity of open benchmark datasets limit existing systems' rapid development and expansion. Most previous works use available small datasets of a maximum of 10 authors for this purpose \cite{das2015experimental, bhai, ourchar}.
    \item Current systems are not scalable and perform poorly for increasing number of authors, especially where number of sample texts per author is limited, which is the case generally for real-life scenarios.
    \item Despite the deep neural network-based model's good performance, they suffer due to the use of small datasets and short length of texts. They typically require many samples for each author, which is costly, time-consuming, and not always available.
    %\item No work has been done to leverage the power of transfer learning in this field as of our knowledge, which can immensely reduce manual labour and increase model re-usability manifold.
\end{enumerate}

% \clearpage

A linguistically driven transfer learning method was proposed in \cite{howard2018universal}, which uses pre-trained language models trained on unsupervised large text corpora to improve performance in various downstream tasks through representation learning. With the help of transfer learning, it is possible to the solve existing problems and build a highly scalable, robust system for authorship attribution. Hence, an elaborate study on the application of transfer learning for author detection in Bangla using language modeling is presented in this paper. The language model acts as a re-usable language learner trained on a large corpus in an unsupervised manner. Then the semantic and syntactic information learned by the language model is leveraged to perform authorship attribution in the target datasets.

\begin{figure}[H]
    \centering
    \includegraphics[width=1\linewidth]{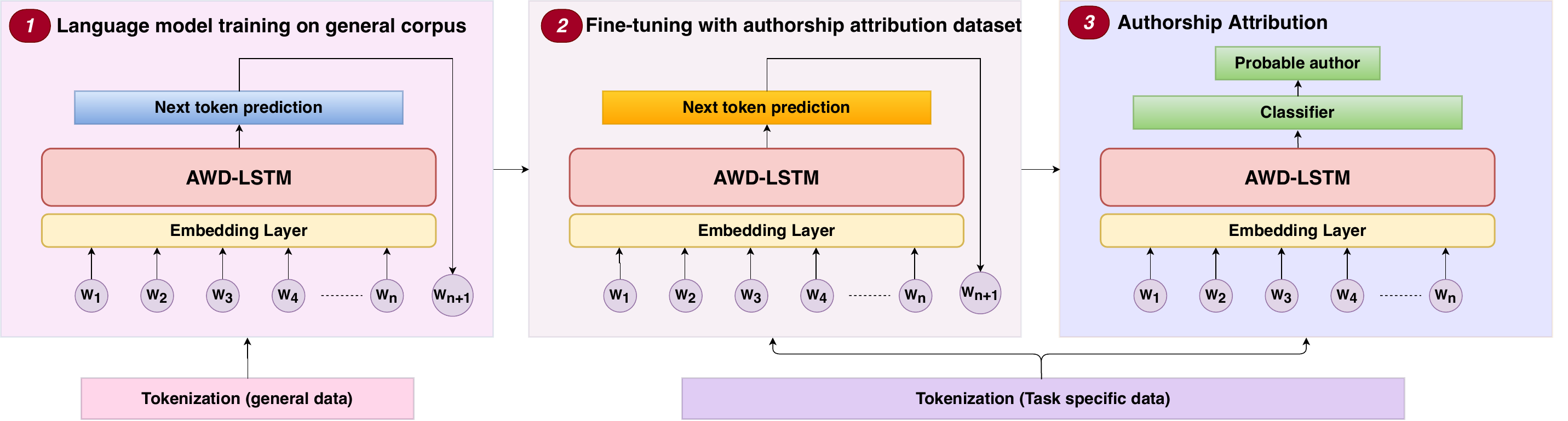}
    \caption{Schematic diagram of the proposed system. The first two steps are language modelling tasks, and the last step is authorship attribution. The embedding layer and the AWD-LSTM base remain the same in all steps, only the final dense layer changes. The trained weights of the fixed parts are passed on from pre-training to fine-tuning to the classification step, updated in each step.}
    \label{overview_methodology}
\end{figure}

In this paper, we present an effective transfer learning approach that uses task-specific optimization with language models pre-trained on two different datasets during the training phase and fine-tuned on target datasets. Figure \ref{overview_methodology} shows a brief overview of our methodology where we use AWD-LSTM (\textbf{A}verage Stochastic Gradient Descent \textbf{W}eight-\textbf{D}ropped \textbf{LSTM}) \cite{AWD} architecture as the language model. The language model is first trained with a general Bangla corpus and then fine-tuned with authorship attribution text in an unsupervised manner. Later, a classifier is added to the pre-trained language models, and supervised authorship attribution is carried out. Additionally, the effects of various tokenization on proposed models are analyzed in terms of performance; the robustness is tested against the existing models by increasing the number of authors and decreasing samples per author. Extensive experiments were done with all the variations, and the best model for AABL was identified. The results demonstrate a clear superiority of the transfer-learning-based approach against the other traditional models.

Our contributions are three-fold:

\begin{enumerate}
    \item In this article, we introduce the largest and most varied dataset for AABL with long text samples of 16 authors in an imbalanced manner, imitating real-world scenarios more closely.

    \item We present an intuitively simple but computationally effective transfer learning approach in which the model is pre-trained on a large corpus, fine-tuned with the target dataset, and later the classifier is trained with labelled data for AABL. To the best of our knowledge, no work has been done to leverage the power of transfer learning for AABL, reducing manual labour and enhancing model re-usability manifold. Experimental results show that the proposed model considerably outperforms the existing models and achieves state-of-the-art performance, efficiently solving the limitations of the previous works. 
    
    \item The various language models trained in this work can be used as pre-trained models for many downstream tasks in Bangla language. All of these pre-trained models, along with the code and dataset of this paper have been released for public use\footnote{\url{https://github.com/tanny411/Authorship-Attribution-using-Transfer-Learning}}.
\end{enumerate}

The paper is organized as follows. Section \ref{related} gives an overview of the previous related work. Section \ref{corpus} describes all the datasets used in this work with brief statistical reports. Section \ref{methodology} contains the methodology of our work. In Section \ref{experiments} we describe the proposed model along with some traditional models used for comparison. Section \ref{results} shows the results and their analysis. Section \ref{pretrained} contains information about the released pre-trained models. Finally, conclusions and directions for further research are summarized in Section \ref{conslusion}.

\section{Related Work}
\label{related}
\subsection{On Authorship Attribution}
Authorship attribution task is an essential research topic and has been prevalent for quite a long time. Research in authorship attribution relies on detecting authors' style through stylometric analysis, assuming that the authors subconsciously use homological idiolect in their writings. Different kinds of feature extraction based methods are implemented to detect these patterns. The features are classified as lexical, character, syntactic, and semantic. Function words, punctuation, ratio of certain words, average word length, average sentence length, vocabulary richness, bag of n-grams, etc., are extracted to determine the authorial styles \cite{sari2018topic, das2015experimental, hossain2017stylometric, el2020using}. Function words are parts of speech that do not have lexical meaning but express grammatical relationships among other words and can be used to identify the authors from their sub-conscious pattern of the use of these words and other writing characteristics. Naive Bayes and Support Vector Machines were used with features such as ratio of punctuation, apostrophes per word, uppercase ratio, etc. \cite{barry2012stylometry}. Word frequency, part-of-speech bigram model, and preterminal tag bigram model were also used \cite{goldmanusing}. Topic modeling approaches were employed to detect authors of unknown texts \cite{seroussi2014authorship, phani2017supervised}

Most of the earlier researchers dealt with small collections where each author may have been heavily inclined towards particular topics which made the authorship attribution task border on topic classification \cite{sari2018topic}. Character n-grams can pick up author nuances combining lexical and syntactic information, and so are an essential set of features \cite{sari2018topic, stamatatos2009survey}. A combination of lexical and syntactic features provides valuable information and has shown enhancement in performance \cite{kreutz2018exploring}. Such combinations can be crucial to improve performance in cross-topic authorship attribution and single-domain attribution \cite{sundararajan2018represents}.

Proceeding with the advancement of deep learning, a large body of work is available on authorship attribution and stylometry using various deep learning models. For example, multi-headed recurrent neural network character language model was used that outperformed the other methods in PAN 2015 \cite{bagnall2016authorship}. Some works used syntactic recurrent neural networks, which learn document representations from parts-of-speech tags and then use attention mechanisms to detect the authorial writing style \cite{jafariakinabad2019syntactic}. Convolutional neural networks have also been employed for this task. Impressive performance was achieved by using character-level and multi-channel CNN for large-scale authorship attribution \cite{ruder2016character}. Character n-grams help identify the authors of tweets, and CNN architectures capture the character-level interactions, representing patterns in higher levels, thus detecting distinct styles of authors \cite{shrestha2017convolutional}. A combination of pre-trained word vectors with one hot encoded POS tag was also used \cite{hitschler2017authorship}. Others investigate syntactic information in authorship task by building separate language models for each author using part-of-speech tags besides word and character-level information \cite{fourkioti2019language}.

\subsection{On Bangla}

Despite the significant progress in English and other western languages for authorship attribution, not much has been done in the Bangla language, especially with transfer learning methods. A notable body of literature exists on the use of hand-drawn features to extract authorial styles from texts. Features such as POS tag count, word frequency, word and sentence length, type-token ratio, unique words percentage, common words, spelling variation, etc., are often used. Some of the such extracted features were used with a dataset of 4 \cite{das2015experimental} and 3 \cite{chakraborty2012authorship} current Bangladeshi authors. The use of N-gram features is the most common along with other extracted features and has been used with probabilistic classification methods \cite{das2011author}, voting classifier with cosine similarity measurements for 6 authors \cite{hossain2017stylometric}, Naive Bayes for 3 authors \cite{anisuzzaman2018authorship, phani2017supervised}, Random forests for 10 authors \cite{islam2017automatic}, and Feed-forward neural networks for 3, 5 and 23 authors \cite{phani2016machine, islam2018authorship, ahmad2020empirical}. SVM was employed on a dataset of 6 authors \cite{pal2017machine}. Islam et al. used unigram, bigram, trigram, part-of-speech like pronoun and conjunction as features, removed stop words, and used Information Gain (IG) to select the most important features. The classification was compared between Naive Bayes, Decision tree, and Random forest classifiers \cite{islam2017automatic}. On the other hand, Phani et al. used term frequency (tf), term frequency-inverse document frequency (tf-idf), and binary presence methods on stop words, word and character unigrams, bigrams, and trigrams to perform author attribution with Naive Bayes, SVM and Decision Trees \cite{phani2017supervised}.

Traditional methods of using n-grams counts consider each feature independent of meaning and context. Instead, all words can be combined under a continuous vector space by representing each word or token with a high dimensional vector. These vectors are called word embedding and can better represent semantic and syntactic similarity among words such that similar words are closer in the vector space than words that are semantically different. The effects of various kinds of word embeddings on authorship attribution were analyzed on a 6 author dataset with multiple neural network architectures showing that using fastText skip-gram word embeddings with a Convolutional Neural Network (CNN) performs best to extract authorial information from texts \cite{bhai, chowdhury2019authorship, chowdhury2019continuous}. Besides, a character-level CNN model was used for 6 to 14 author detection where character embeddings were pre-trained on a larger corpus \cite{ourchar}.

\subsection{On Transfer Learning}

Transfer learning is the process of reusing a model trained on an initial task as a starting point for a different task. Deep learning approaches have been using this method to boost model performance and save an enormous amount of time on various computer vision tasks and natural language processing (NLP). Transfer learning has been prevalent in computer vision for a long time, and pre-trained models on massive datasets are readily available. Natural language processing started approaching transfer learning by using pre-trained word embeddings which aim at only the first layer of the model. Some approaches combine multiple derived embeddings at different layers \cite{peters2018deep}. The idea of using transfer learning from language models has been approached \cite{dai2015semi} but not widely adopted due to its need for large-scale datasets. To address these issues, a method called ULMFiT was proposed that successfully enables robust transfer learning in any NLP task \cite{howard2018universal}. Later transformer-based transfer learning approaches were able to reduce training time significantly \cite{devlin2018bert, liu2019roberta, radford2018improving}. These architectures use attention mechanism, which works on the entire sequence simultaneously instead of working token by token like in LSTM. Many variations of these models have been used in numerous NLP tasks. Pre-trained models using multiple languages are also made available for various transformer-based architectures \cite{devlin2018bert, lample2019cross, yang2019xlnet, conneau2019unsupervised}.

Transfer learning has been adopted in Bangla language tasks in both computer vision and NLP. Various computer vision tasks exist such as handwritten Bangla word recognition \cite{pramanik2020segmentation}, Bangla sign language recognition \cite{nishat2019unsupervised}, Bengali Ethnicity and Gender Classification \cite{jewel2019bengali} etc. In NLP, transfer learning started with word, and character embeddings \cite{bhai, ourchar}. BengFastText was created as the largest Bangla word embedding model using 250 million articles \cite{karim2020classification}. BERT \cite{devlin2018bert} was used for sentiment analysis in code-mixed English-Bangla social media data \cite{jamatia2020deep}. The contextualized embeddings from BERT were used for Bangla named entity recognition task \cite{ashrafi2020banner}. Transfer learning from pre-trained models was also used for detecting fake news in the Bangla \cite{hossain2020banfakenews}. No such work on authorship attribution in the Bangla language has been done to date.

\section{Corpora}
\label{corpus}
%https://colab.research.google.com/drive/1wKO3g4WxGcSZwRibaOywSX3B_hlI0ozw
The scarcity of standard benchmark datasets in the Bangla language makes the task of authorship attribution quite tricky. Most researchers in Bangla have used datasets containing 3-6 authors on average. For this purpose, we create a dataset of 16 authors and name it BAAD16. It is the largest dataset for authorship attribution in Bangla, containing texts from 16 authors and 750 words per sample with 17966 samples in total (More description in Section \ref{aa}), unlike most datasets which have up to 10 authors at most \cite{phani2016machine, das2015experimental, islam2018authorship, islam2017automatic}. One exception is a dataset with 23 authors \cite{ahmad2020empirical}, it is not publicly available for use, and the word count for each document in this dataset is only 50 words on average, making it a short-text detection dataset. We also use a dataset of 6 authors called BAAD6, which has 1000 words per document. Both of these datasets are long-text-containing datasets, which becomes more challenging for traditional machine learning techniques to handle. A comparison of existing datasets, along with their respective performance, will be shown in Section \ref{results} Table \ref{dataset_comparison}. Besides, a larger corpus of Bangla text is required to perform pre-training in an unsupervised manner. For this step, we use a News corpus and a Wikipedia corpus. In total, four datasets were used in different steps of the experiments. Descriptions of all the datasets used in this work are given below.

\subsection{BAAD16: Bangla Authorship Attribution Dataset of 16 authors}
\label{aa}
\begin{figure}[H]
\begin{subfigure}{.495\textwidth}
\centering
\includegraphics[width=1\linewidth]{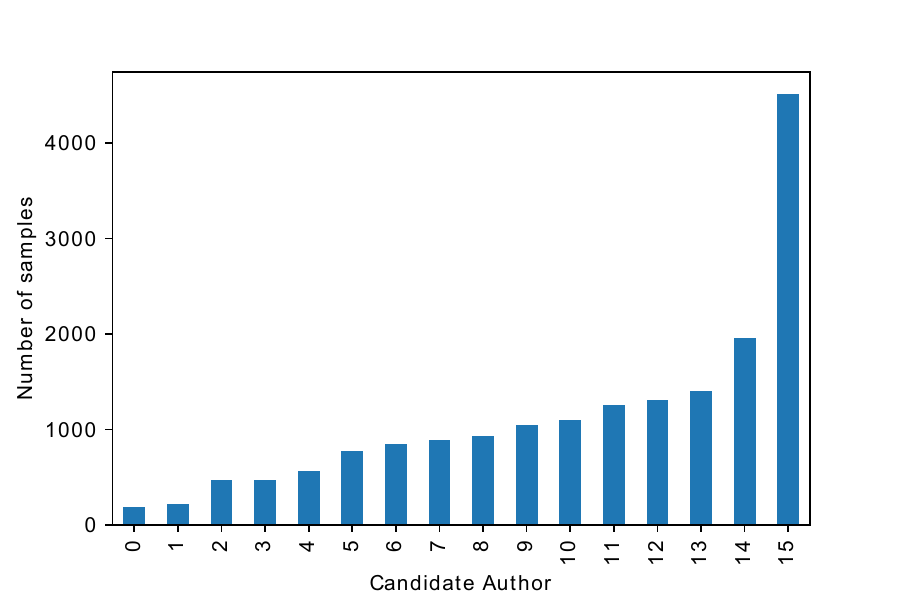}  
\caption{Sample distribution}
\label{sample_dist_16}
\end{subfigure}
% \end{figure}
% \begin{figure}[H]
% \ContinuedFloat
\begin{subfigure}{.495\textwidth}
\centering
\includegraphics[width=1\linewidth]{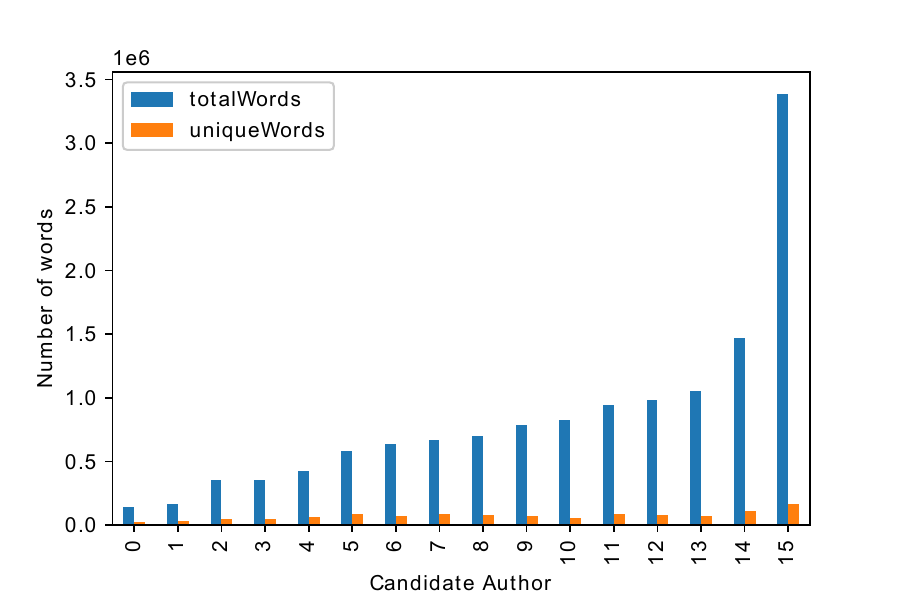}  
\caption{Words distribution}
\label{word_dist_16}
\end{subfigure}
\caption{BAAD16 dataset: Distribution per author. Authors are indicated by their index number. See indices in Table \ref{ourcorpustable}}
\label{ourdatasetfig}
\end{figure}

This dataset has been created by scraping text from an online Bangla e-library using a custom web crawler and contains literary works of different Bangla famous writers. It contains novels, stories, series, and other works of 16 authors. Each sample document is created with 750 words. The aim is to create a long-text dataset, but the length is often capped by the memory usage of long input strings and the length of content usable by models to less than 1000 words. Therefore we chose a length of 750 to help provide as much context as possible for any given piece of text. In Table \ref{ourcorpustable} details are shown about the dataset, and Figure \ref{sample_dist_16} shows the distribution of text samples for each of the 16 authors. This dataset is imbalanced, as apparent from Figure \ref{ourdatasetfig} and resembles real-world scenarios more closely, where not all the authors will have a large number of sample texts. This is the largest dataset in Bangla literature for authorship attribution in terms of the number of words available per author and the total number of words (13.4+ million). Figure \ref{word_dist_16} shows the distribution of the total number of words and the unique word count for each author. We further create multiple balanced subsets of this dataset for various numbers of authors in order to measure model stability with increasing author numbers. In this case, the datasets are truncated to the minimum number of samples available per author.

\begin{table}[h]
\caption{BAAD16 dataset statistics}
\begin{tabular}{|c|l|c|c|c|}
\hline
\textbf{Author Id} & \textbf{Author Name} & \textbf{Samples} & \textbf{Word Count} & \textbf{Unique Word} \\ \hline
0                  & zahir\_rayhan        & 185              & 138k                & 20k                  \\ \hline
1                  & nazrul               & 223              & 167k                & 33k                  \\ \hline
2                  & manik\_bandhopaddhay & 469              & 351k                & 44k                  \\ \hline
3                  & nihar\_ronjon\_gupta & 476              & 357k                & 43k                  \\ \hline
4                  & bongkim              & 562              & 421k                & 62k                  \\ \hline
5                  & tarashonkor          & 775              & 581k                & 84k                  \\ \hline
6                  & shottojit\_roy       & 849              & 636k                & 67k                  \\ \hline
7                  & shordindu            & 888              & 666k                & 84k                  \\ \hline
8                  & toslima\_nasrin      & 931              & 698k                & 76k                  \\ \hline
9                  & shirshendu           & 1048             & 786k                & 69k                  \\ \hline
10                 & zafar\_iqbal         & 1100             & 825k                & 53k                  \\ \hline
11                 & robindronath         & 1259             & 944k                & 89k                  \\ \hline
12                 & shorotchandra        & 1312             & 984k                & 78k                  \\ \hline
13                 & shomresh             & 1408             & 1056k               & 69k                  \\ \hline
14                 & shunil\_gongopaddhay & 1963             & 1472k               & 109k                 \\ \hline
15                 & humayun\_ahmed       & 4518             & 3388k               & 161k                 \\ \hline
\multicolumn{2}{|c|}{\textbf{Total}}      & 17966            & 13474500            & 590660               \\ \hline
\multicolumn{2}{|c|}{\textbf{Average}}    & 1122.875         & 842156.25           & 71822.25             \\ \hline
\end{tabular}
\label{ourcorpustable}
\end{table}

\subsection{BAAD6: Bangla Authorship Attribution Dataset of 6 authors}
\label{aa2}
BAAD6 is a 6 Author dataset collected and analyzed by Hemayet et al. \cite{bhai}.
%\cite{al2021identifying}
This dataset's total number of words and unique words are 2,304,338 and 230,075 respectively. The data was obtained from different online posts and blogs. This dataset is balanced among the 6 Authors with 350 sample texts per author. The total word count and number of unique words per author are shown in Table \ref{bhaicorpustable}. This is a relatively small dataset but is noisy given the sources it was collected from and its cleaning procedure. Nonetheless, it may help evaluate the proposed system as it resembles texts often available on the Internet.

\begin{table}[h!]
\centering
\caption{BAAD6 dataset statistics}
\begin{tabular}{|c|c|c|c|} 
\hline
\textbf{Author} & \textbf{Samples} & \textbf{Word count} & \textbf{Unique word}\\
\hline
fe & 350 & 357k & 53k\\
ij & 350 & 391k & 72k\\
mk & 350 & 377k & 47k\\
rn & 350 & 231k & 50k\\
hm & 350 & 555k & 72k\\
rg & 350 & 391k & 58k\\
\hline
\textbf{Total} & 2100 & 2304338 & 230075\\
\hline
\textbf{Average} & 350 & 384056.33 & 59006.67\\
\hline

\end{tabular}
\break
\label{bhaicorpustable}
\end{table}

% \vspace{10pt}

\subsection{News Corpus}
\label{news}
This corpus comprises of various Bangla newspaper articles in 12 different categories. It is significantly large compared to the datasets used in previous works in Bangla. There are 28.5+ million word tokens in this corpus, and the number of unique words is 836,509 forming around 3\% of the total vocabulary. Table \ref{corpustablenews} shows the details of the dataset, along with some necessary statistical measures.

\begin{table}[h!]
\centering
\caption{News dataset statistics}
\begin{tabular}{|c|c|c|c|} 
\hline
\textbf{Category} & \textbf{Samples} & \textbf{Word count} & \textbf{Unique word}\\
\hline
opinion	& 8098 & 4185k & 243k\\
international & 5155 & 1089k & 86k\\
economics & 3449 & 909k & 58k\\
art & 2665 & 1312k & 154k\\
science & 2906 & 697k & 76k\\
politics & 20050 & 6167k & 196k\\
crime & 8655 & 2016k & 128k\\
education & 12212 & 3963k & 225k\\
sports & 11903 & 3087k & 174k\\
accident & 6328 & 1086k & 77k\\
environment & 4313 & 1347k & 103k\\
entertainment & 10121 & 2669k & 204k\\
\hline
\textbf{Average} & 7988 & 2377803 & 144342\\
\hline
\textbf{Total} & 95855 & 28533646 & 836509\\
\hline
\end{tabular}
\label{corpustablenews}
\end{table}

\subsection{Wikipedia Corpus}
%10/june/2019
\label{wiki}
%bnwiki-latest-pages-articles-multistream.xml.bz2
Besides the News corpus, we also test our method by pre-training on a Wikipedia corpus. A subset of the Bangla Wikipedia text was collected from the Bangla wiki-dump\footnote{collected on 10th June 2019}. The files are then merged, and each article is selected as a sample text. All HTML tags are removed, and the page's title is stripped from the beginning of the text. This dataset contains 70,377 samples with the total number of words approximating at 18+ million. The entire dataset has more than 1.2 million unique words, which is 7\% of the total word count. Compared to the news dataset, this makes the Wikipedia dataset more varied regarding the types of words used.
%1,289,249

\section{Methodology}
\label{methodology}
\subsection{Overview}
Transfer learning generally implies three basic steps: (i) Pre-training, (ii) Fine-tuning, and (iii) Downstream tasks. In this case, the downstream task is authorship attribution. We pre-train the models from scratch due to the unavailability of pre-trained Bangla language models. First, we tokenize our texts in one of three ways - word, subword, or character. Then we train the language model on a large Bangla text corpus. To perform the downstream task, we fine-tune the language model using text from the tokenized authorship attribution dataset in an unsupervised manner and then finally perform classification by adding a classifier to the trained language model. The entire procedure, starting from tokenization to author identification, was previously shown schematically in Figure \ref{overview_methodology}.

The idea of using such a setup is much like the traditional approach of using a single word-embedding layer, only here, the language model is a multi-layered deep neural network on its own, which is capable of holding much more information in its weights than a single layer of embedding matrix could \cite{howard2018universal}. The traditional approaches of authorship attribution in Bangla Language, even in deep learning methods, implement a classification model with word representations or other extracted features, taking minimal regard for the language on its own. However, the general idea of this research is to capture the essence of the Bangla language first, then proceed to the more granular task of classification with the help of language models. After the training phase, we test with some examples and find that the models created in our experiments can complete entire Bangla paragraphs somewhat meaningfully with minimal grammatical or semantic flaws. So, it is safe to assume that the weights of the language models hold enough relevant information to understand the patterns at which Bangla words appear one after another instead of a single word-embedding layer, which only represents the relationship among words.

\subsection{Tokenization}
\label{tokenization}
\begin{figure}[h]
    \centering
    \includegraphics[width=0.8\linewidth]{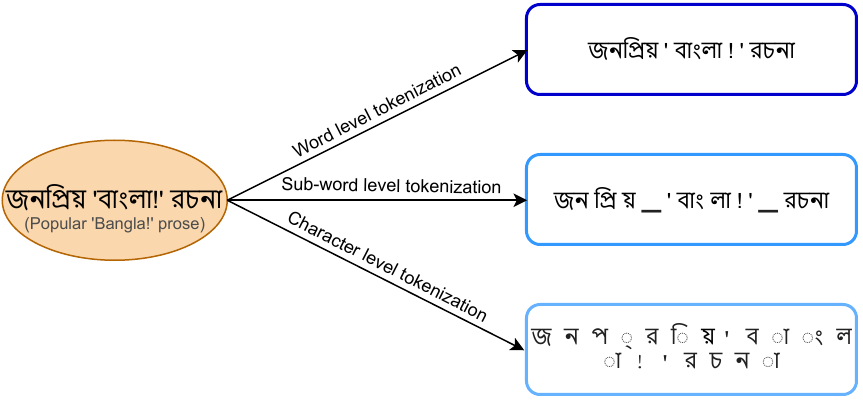}
    \caption{Example of three kinds of tokenization of a sample text. Punctuation added for demonstration purposes.}
    \label{tokenization_sample}
    \end{figure}

Traditionally tokenization in Bangla has been carried out at the word level, separating words by spaces, including punctuation or special characters as separate tokens. Nevertheless, Bangla language has some distinct characteristics which make a mere separation by words less meaningful. Lemmatization and stemming provide a way for removing inflection, but this process remains arduous, and only basic rule-based systems are available to perform this task in Bangla. Besides, the removed parts of the words are not always wholly meaningless and may provide meaning in terms of gender, person, case, number, or animacy. Information provided by declension is also a necessary part of the language to consider. Therefore, besides word-level basic tokenization, we also perform sub-word and character level tokenization and compare the various methods' effects in the final result. All these methods are described elaborately in the following subsections, and the tokenization of a sample text is shown in Figure \ref{tokenization_sample}.
    
    \subsubsection{Word level}
   Word level tokenization is performed mainly by considering words separated by spaces as separate tokens. Besides, words that occur less than three times are discarded, considering them as rare words, names, or misspellings. We select the most frequent 60,000 words as the vocabulary to proceed to the network. Unknown words are replaced with the $<$unk$>$ token. Some specialized tokens are also added, such as the beginning of a string $<$bos$>$, end of a string $<$end$>$, padding $<$pad$>$, character repetition, and word repetition representing tokens, etc.
   
   \clearpage
   
    \subsubsection{Sub-word level}
    
    Sub-word tokenization means dividing a word into parts and using each part as a token. Despite having considerable inflection, there is no specialized sub-word tokenizer for Bangla language. To tackle this problem, sub-word tokenization was performed using SentencePiece tokenizer \cite{kudo2018sentencepiece}. SentencePiece is a language independent tokenizer that tokenizes raw sentences in an unsupervised manner. Using SentencePiece, the unigram segmentation algorithm \cite{kudo2018subword} was employed to create the sub-word vocabulary. For training, we chose 30,000 most frequent tokens \cite{howardsubword}. Tokens appearing less than three times were discarded and replaced with $<$unk$>$ token. Other tokens include $<$s$>$ as the start and $<$/s$>$ as the end of a sentence.
    
    \subsubsection{Character level}
   
    In character level tokenization, a sentence is split into characters, and each character is considered a token. In this case, the total number of tokens is much smaller, 188 in our case, by combining Bangla, English alpha-numerals, and special characters. The language model generates one character at a time, concatenated to form words,  sentences, and even paragraphs. Besides the vocabulary, we also use special tokens, just as was used in word-level tokenization. The use of character-level tokenization not only reduces the vocabulary size drastically but also removes the bottleneck for out-of-vocabulary words, misspellings, etc. Characters can be used to build correlation among groups of characters despite the inflection and declension that occur and thus add additional meaning to the word. Related words can be kept as they are, without discarding any part and losing any information, and still be recognized as being related.

    \begin{figure}[h]
    \begin{subfigure}[t]{.49\textwidth}
    \centering
    \includegraphics[width=1\linewidth]{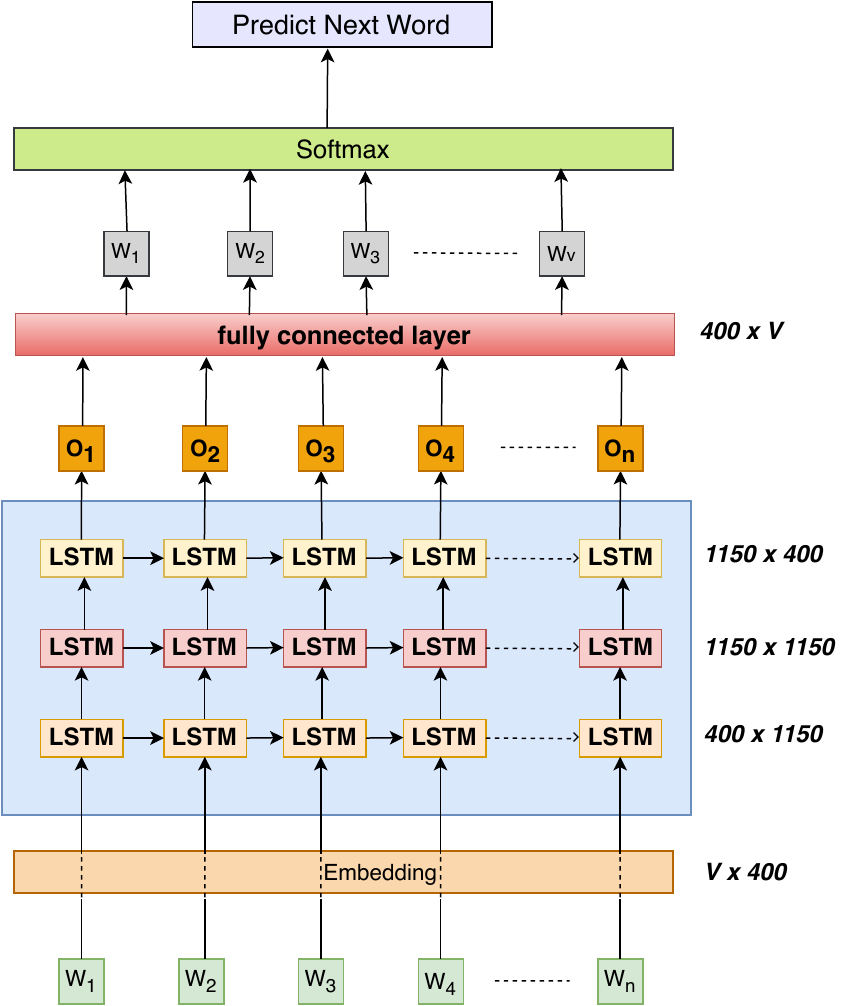}
    \caption{Language Model Architecture}
    \label{awdlstm_lm}
    \end{subfigure}
    \begin{subfigure}[t]{.49\textwidth}
    \centering
    \includegraphics[width=1\linewidth]{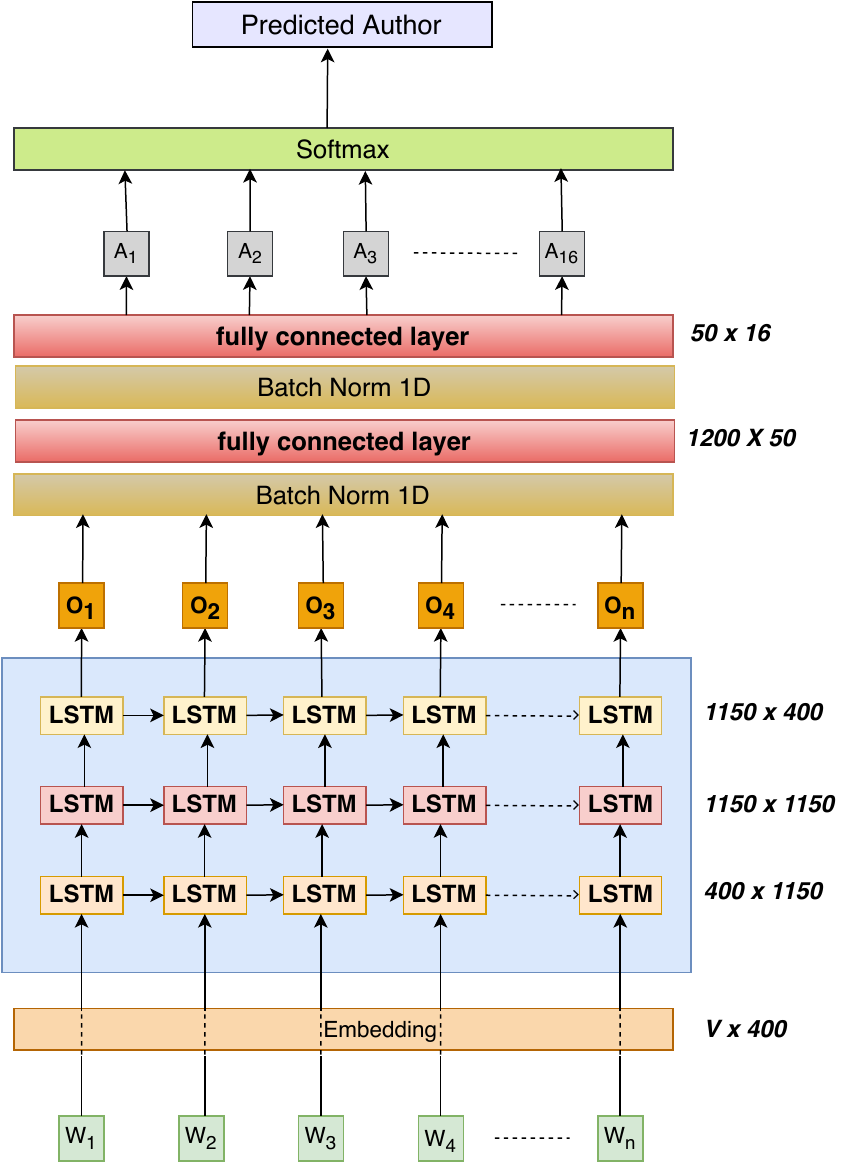}  
    \caption{Classification Model Architecture}
    \label{awdlstm_cls}
    \end{subfigure}
    
    \caption{Simplified diagram of the architectures. The base (embedding and LSTM layers) of the model remain the same to transfer learned weights, but the classifier parts are changed when the tasks are changed.}
    \label{architectures}
    \end{figure}

\subsection{Proposed Architecture}

    \subsubsection{Language Model}
    The patterns in the writings of authors recur, and this nature of the task suggests the use of a recurrent neural network (RNN). A simplified diagram of the architecture is shown in Figure \ref{awdlstm_lm}. The architecture consists of an encoder and a decoder part. The encoder network starts with an Embedding layer with an embedding size of 400, followed by 3 regular LSTM layers, each with 1150 hidden nodes. It has a few short-cut connections and numerous drop-out hyper-parameters but does not use any attention mechanism. The decoder is formed by a dense and a softmax linear layer that provides the probabilistic estimations for the next word over the vocabulary. Input passes through the encoder, and the outputs from the LSTMs are passed on to the decoder, where the prediction is generated. We used Adam optimizer \cite{adam} and flattened cross-entropy loss function to train our models.

    % \clearpage
    
\begin{figure}[H]
    \centering
    \vspace{20pt}
    \includegraphics[scale=0.36]{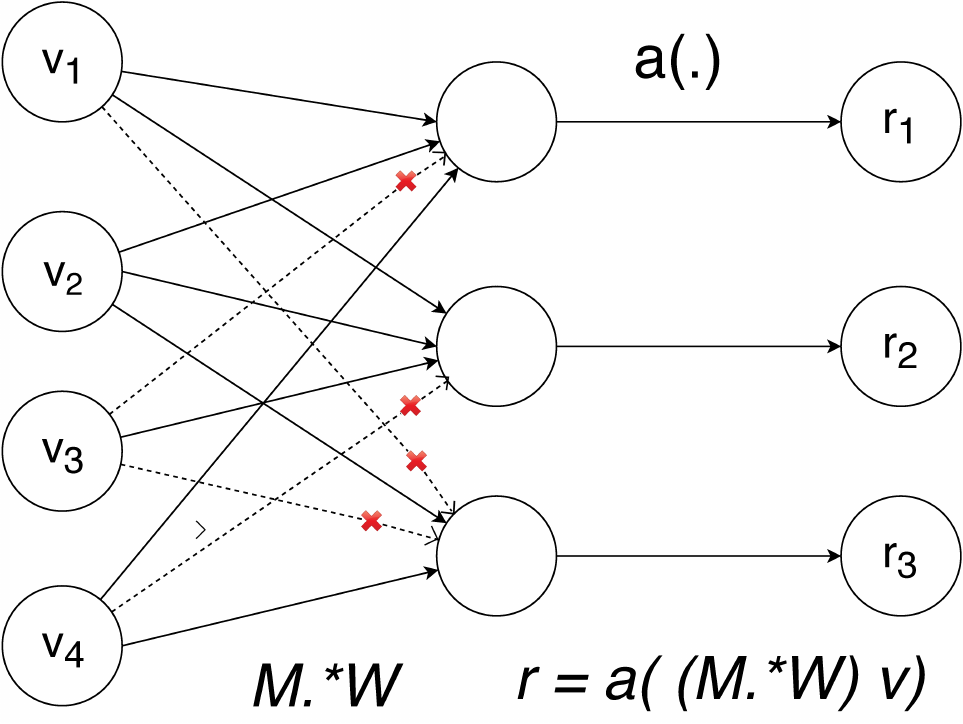}
    % \caption{DropConnect Network\protect\footnote{\url{https://cds.nyu.edu/projects/regularization-neural-networks-using-dropconnect/}}}
    \caption{DropConnect Network \cite{dropconnect_image_cite}}
    \label{Fig: DropConnect}
    \end{figure}

    In this case, we employed a particular variant of the LSTM called the AWD-LSTM \cite{merity}. It stands for ASGD(Average Stochastic Gradient Descent) Weight-Dropped LSTM. This model provides special regularization techniques such as drop-connect \cite{wan2013regularization} and other optimizations that make it a suitable choice for generalizing context and language modelling. In traditional models, overfitting is a significant issue. Drop-connect handles this issue by randomly selecting the activation subset on the hidden-to-hidden weight matrices. This preserves the LSTMs ability to remember long-term dependency yet not overfit. Figure \ref{Fig: DropConnect} provides a graphical example of the drop-connect network.
    
    %% footnote does not work
    
    Among some other techniques is the generation of a dropout mask on the first call, namely the variational dropout \cite{variable_dropout}. Embedding Dropout \cite{variable_dropout}  and reduction of embedding size helped the architecture in achieving high performance in language modelling \cite{AWD}. This model is therefore chosen for the successful application of transfer learning \cite{howard2018universal}.
    
    \subsubsection{Classifier}
    \label{classifier}
    The classifier is built on top of the language model by only changing the decoder part of the model. Two linear layers are added with batch normalization, and dropout \cite{howard2018universal}. Activation function ReLU is used for intermediate layers and softmax for the output layer, which outputs the probability distribution among the given authors. The author with the highest probability value is then selected as the predicted author. A diagram of the classifier architecture is shown in Figure \ref{awdlstm_cls}.
    
\subsection{Training techniques} 
In recent research, training neural networks efficiently has become a cumulative task of proper architecture selection, suitable techniques application, and fine-tuning. This section briefs the techniques used in this paper.

\subsubsection{1cycle policy}
\begin{figure}[H]
\captionsetup{justification=centering}

\begin{subfigure}{.325\textwidth}
  \centering
  \includegraphics[width=1\linewidth]{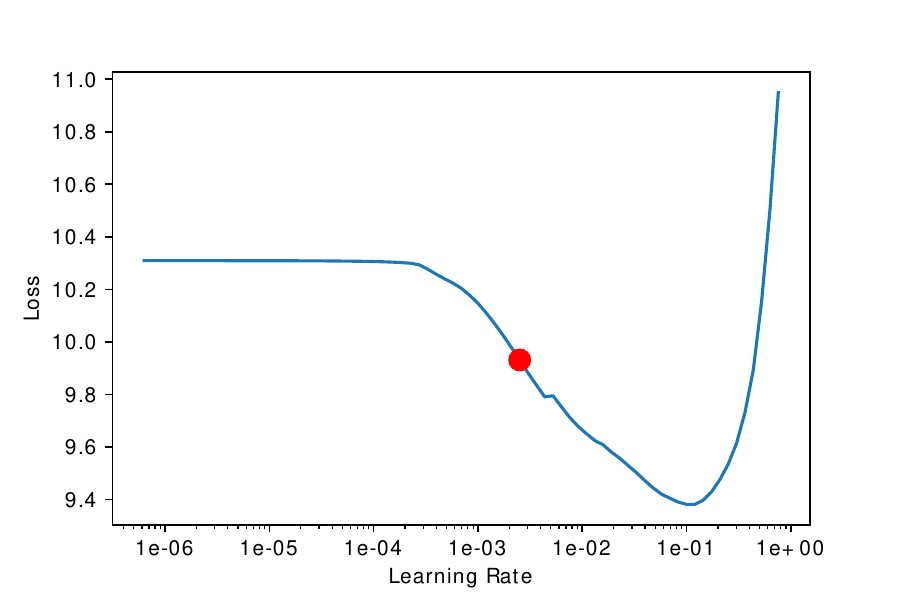}  
  \caption{Cyclical learning rate plot achieved\\ during the pre-training phase.}
%   \label{fig:sub-first}
\end{subfigure}
% \end{figure}
% \begin{figure}[h]
% \ContinuedFloat
\begin{subfigure}{.325\textwidth}
  \centering
  \includegraphics[width=1\linewidth]{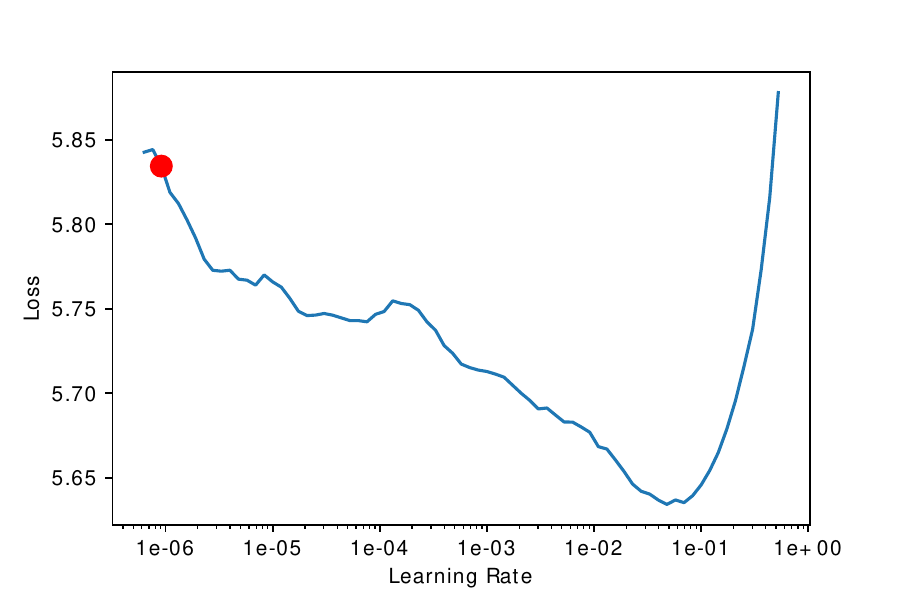}  
  \caption{Cyclical learning rate plot achieved\\ during the fine-tuning phase.}
%   \label{fig:sub-second}
\end{subfigure}
% \end{figure}
% \begin{figure}[H]
% \ContinuedFloat
\begin{subfigure}{.325\textwidth}
  \centering
  \captionsetup{justification=centering}
  \includegraphics[width=1\linewidth]{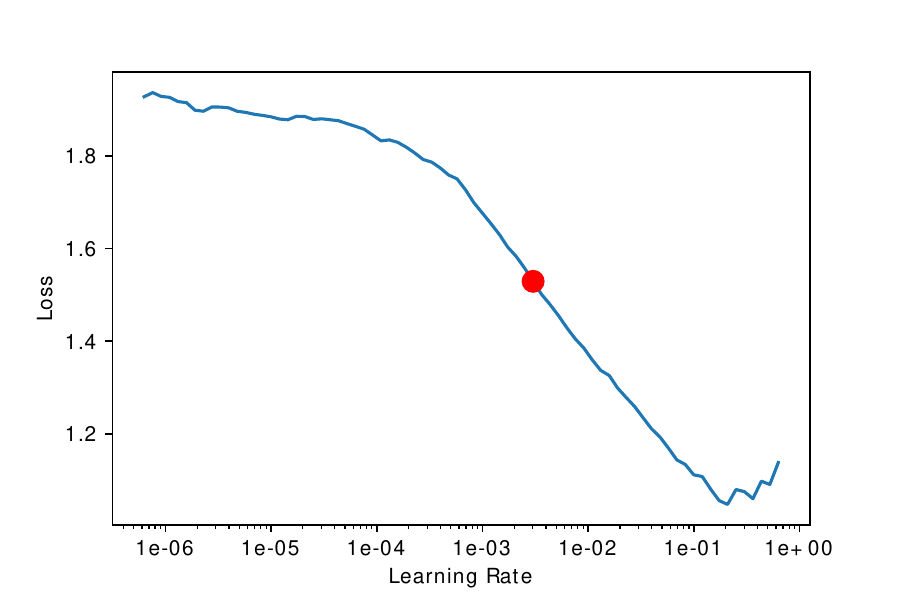}  
  \caption{Cyclical learning rate plot achieved\\  during the classification phase.}
%   \label{fig:sub-third}
\end{subfigure}
\caption{The cyclical learning rate for various stages of training. The entire dataset is swept once, and losses for various learning rates(in log scale) are plotted as the blue line. The red points show the suggested learning rate, where the loss has a negative slope and continues as such for a while.}
\label{clr}
\end{figure}

The learning rate is one of the most crucial hyper-parameter that drastically affects experiments and has been manually tuned for quite some time. A unique approach was published in this regard \cite{clr}. In this method, a single trial is run over the dataset, starting with a low learning rate and is increased exponentially batch-wise. The loss observed for each value of the learning rate is recorded. The point of decreasing loss and a high learning rate is selected as the optimal value of the learning rate. Examples of learning rate finder diagrams obtained in our different experimental steps are shown in Figure \ref{clr}.

% \vspace{10pt}

\subsubsection{Discriminative learning rate}
\begin{figure}[H]
\includegraphics[scale=0.45]{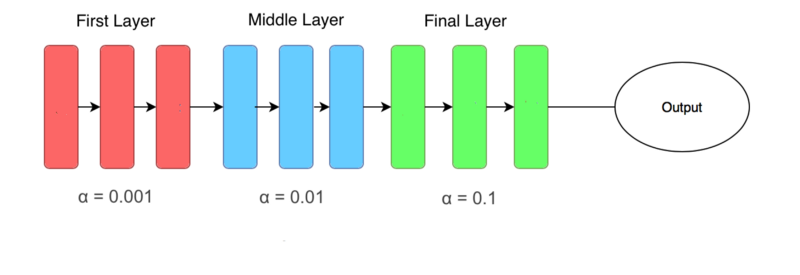}
\centering
\caption{Application of Discriminative Learning Rates \cite{tenthings}. The initial layers (red) have a lower learning rate to retain the pre-trained weights, while later layers (green) have a higher learning rate to tune faster to the task at hand.}
\label{diff_learn}
\end{figure}

The chosen learning rate is often fixed throughout the training process and among all the layers. The various layers in a deep learning model contain different levels of information, and sometimes it may be necessary to allow for some information to be preserved or changed slowly or fast compared to the others. A technique known as discriminative learning rate \cite{howard2018universal} solves the problem. This method ensures that the later layers of a neural network train faster than the base layers by applying a range of learning rates across different layers. Building deep learning models on top of the embedding layers has shown promising results in the recent past, which implies that the later layers of the model need to be modified. This can be accomplished using a higher learning rate for the last layers, whereas for the base layers, we use a low learning rate to slowly change the pre-trained weights and retain the most learned information. Figure \ref{diff_learn} illustrates a generic deep learning architecture to show discriminative learning rates.

% \clearpage

\subsubsection{Gradual unfreezing}
When a task-specific model is trained on top of a pre-trained model in transfer learning, all the layers' weights are altered together. This method carries noise back to the base layers from the newly attached (randomly initialized) layers. To prevent such abrupt alteration and, therefore, catastrophic forgetting, we employ gradual unfreezing \cite{howard2018universal}. This method first freezes the initial layers and trains only the last layers; then, the pre-trained layers are unfrozen one by one and trained further to tune to the problem-specific domain. Thus, the learned information of the pre-trained model is used efficiently. %%%%%%img?

 \subsubsection{Slanted triangular learning rates}
 The original concept of stochastic gradient descent \cite{SGDR} tries to employ the idea that the neural networks should be getting closer to the loss's global minimum value. As the global minimum for loss gets closer to the minimum, the learning rate should decrease quite obviously. If it doesn't, the system may fall in an infinite loop as it jumps from one side of the global minima to another because it keeps subtracting the multiple of the large learning rate selected. Slanted triangular learning rate (STLR) linearly increases the learning rate to converge quickly and then linearly decays it according to an updated schedule to tune the parameters \cite{howard2018universal}.

 \subsubsection{Stochastic Gradient Descent with Restarts}
\begin{figure}[H]
\includegraphics[scale=0.33]{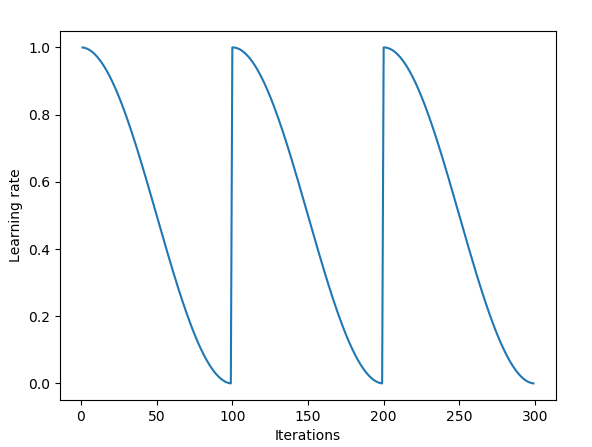}
\centering
\caption{Resetting Learning Rate after each Epoch and reducing gradually by cosine annealing \cite{tenthings}}
\label{sgdr_solution}
\end{figure}

 The neural network training phase will likely encounter local minima in the loss curve rather than the global minimum. The model may get stuck at the local minima instead of reaching for the global minima in such cases. A technique that approaches this problem involves a sudden increase in learning rates hoping that it will 'jump' over the local minima if there are any. The process is called stochastic gradient descent with restarts (SGDR) and was introduced in \cite{SGDR}. SGDR suddenly increases the learning rate and decreases it again by cosine annealing. Figure \ref{sgdr_solution} shows how learning rates are restarted after every epoch to avoid the problem.
 
\subsection{Training}
The primary approach to performing transfer learning with the help of language models closely follows ULMFiT \cite{howard2018universal}. The entire procedure is divided into three broad steps, as shown previously in Figure \ref{overview_methodology}. Each step has been explained in the following subsections.

% \clearpage
    
\begin{figure}[h]
\fbox{\includegraphics[scale=0.9]{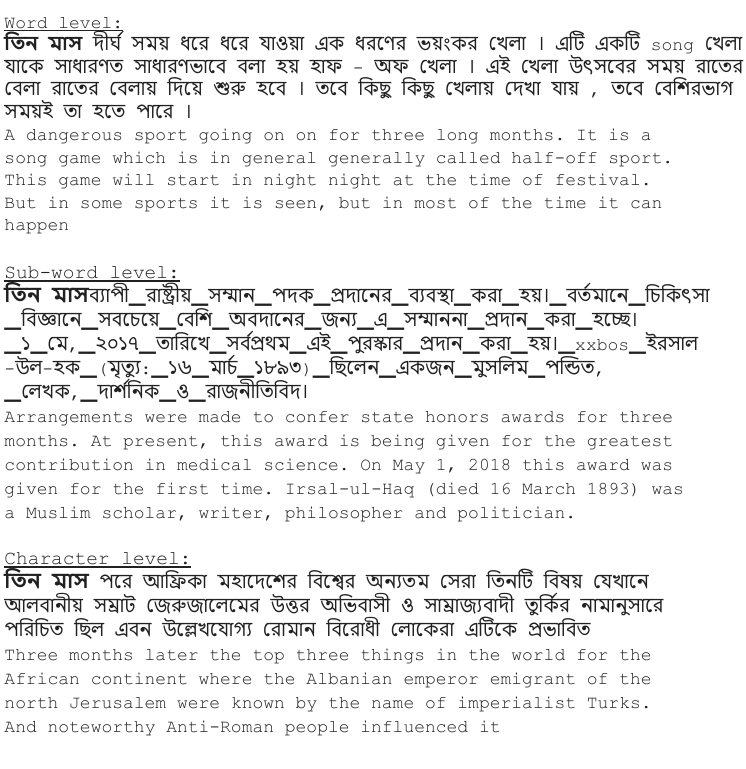}}
\centering
\caption{Text generation from the different trained language models.}
\label{textfig}
\end{figure}

\subsubsection{Language model Pre-training}
Two reasonably large Bangla datasets have been used to perform this training step. The aim is for the model to \emph{learn} the Bangla language through language modeling. For this, we used the Wikipedia dataset and News dataset separately; these contain a variety of Bangla texts that can generalize to most text that may appear in downstream tasks. For all the models, batch size of 32, back-propagation-through-time (bptt) of 70, and weight decay of 0.1 are used on all layers. We used a drop-out multiplier of 0.5 to the ratios selected in \cite{howard2018universal} for all the layers to avoid over-fitting. An appropriate learning rate for each model is selected through a cyclic learning rate finder \cite{clr}. This is depicted in Figure \ref{clr}. Sometimes learning rates needed to be altered as training progressed, as determined by the loss of the validation set. For example, when loss starts to increase, the model is trained from a previous checkpoint with a lower learning rate. This is shown as comma-separated learning rates in Table \ref{ulmhyp}. Each epoch employed stochastic gradient descent with restarts (SGDR)  mentioned in Section \ref{related}. All models are trained until they start to show signs of over-fitting, which is 30 epochs for the word and sub-word levels and 15 epochs for the character-level models. Important hyper-parameters are mentioned in Table \ref{ulmhyp} for all the models. 

\begin{table}[h]
\centering
\caption{Pre-training Language model hyper-parameters}
\begin{tabular}{| l | l | l | l | l |}
\hline
\textbf{Dataset} & \textbf{Tokenization} & \textbf{Epoch} & \textbf{Batch size} & \textbf{Learning rate} \\
\hline

\multirow{3}{*}{News} & Word        & 30 & 32 & 1e-3 \\ \cline{2-5}
                      & Sub-word    & 30 & 32 & 1e-2,1e-3 \\ \cline{2-5}
                      & Character   & 15 & 32 & 1e-2,1e-6,1e-2 \\ \cline{1-5}
\multirow{3}{*}{Wiki} & Word        & 30 & 32 & 1e-2,1e-3 \\ \cline{2-5}
                      & Sub-word    & 30 & 32 & 1e-3 \\ \cline{2-5}
                      & Character   & 15 & 32 & 1e-2 \\ \cline{1-5}

% \hline
\end{tabular}
\label{ulmhyp}
\end{table}

After training, the models begin to learn Bangla. When provided with one or more initial words, the models can complete entire paragraphs. Illustrations of models' prediction are shown in Figure \ref{textfig} from the Wikipedia dataset trained model for different levels of tokenization. All three models were given the starting words 'Three months' in Bangla, and the rest were generated. Note that the outputs were tokens that were bound together, removing extra spaces to form words where required (except in word-level tokenization, where output is already in words). English translations for the generated text are provided under each paragraph. We reproduced all grammatical and semantic errors produced by the model in English as well. Although the text, on the whole, does not make much sense, the model has learned to form reasonable phrases and words. This learning is leveraged in the fine-tuning and classification stages.

% \clearpage

\subsubsection{Language model Fine-tuning}
After training the language model on general text, it is fine-tuned to the task-specific type of text. Each model from the previous section (pre-trained on news and Wikipedia dataset) is then fine-tuned on two authorship attribution datasets as mentioned in Section \ref{corpus}. This step allows the model to get accustomed to the authorial writing style. The target datasets are split into training and testing sets in an 80\%-20\% ratio, and only the training set is used in this phase. The original language model is first loaded and frozen to tune the model except the last layers that hold the most task-specific information. Using cyclic learning rate finder \cite{clr}, the learning rate for each model is determined. The models were trained for two epochs and then trained for two more epochs by unfreezing another layer group, repeating it one more time before unfreezing the entire model. The models are trained as long as the loss keeps decreasing, and training is halted on signs of over-fitting. Learning rates are altered as required by observing the loss of the validation data, depicted by comma-separated values in Table \ref{fithyp}. The rest of the model settings are kept the same as before.
 Batch size, along with other hyper-parameters is summarized in Table \ref{fithyp}.

\begin{table}[h]
\centering
\caption{Fine-Tuning Language model hyper-parameters}
\begin{tabular}{| c | c | c | c | c | c |}
\hline
\textbf{\begin{tabular}[c]{@{}c@{}}Pre-training\\ Dataset\end{tabular}} & \textbf{Tokenization} & \textbf{\begin{tabular}[c]{@{}c@{}}Fine-tuning\\ Dataset\end{tabular}} & \textbf{Epoch} & \textbf{Batch size}& \textbf{Learning rate} \\
\hline

\multirow{6}{*}{News} & \multirow{2}{*}{Word}      & BAAD16 & 19  & 32  & 1e-3 \\ \cline{3-6}
                      &                            & BAAD6 & 10 & 32 & 1e-2   \\ \cline{2-6}
                      & \multirow{2}{*}{Sub-word}  & BAAD16 & 22 & 32 & 1e-3,1e-4 \\ \cline{3-6}
                      &                            & BAAD6 & 10 & 32 & 2e-2 \\ \cline{2-6}
                      & \multirow{2}{*}{Character} & BAAD16 & 14  & 32 & 1e-2 \\ \cline{3-6}
                      &                            & BAAD6 & 10 & 128 & 1e-3 \\ \cline{1-6}
\multirow{6}{*}{Wikipedia} & \multirow{2}{*}{Word} & BAAD16 & 26 & 32 & 1e-2,1e-3,1e-4 \\ \cline{3-6}
                      &                            & BAAD6 & 10 & 32 & 1e-2  \\ \cline{2-6}
                      & \multirow{2}{*}{Sub-word}  & BAAD16 & 24 & 32  &  1e-2,1e-3 \\ \cline{3-6}
                      &                            & BAAD6 & 10 & 32 & 2e-2 \\ \cline{2-6}
                      & \multirow{2}{*}{Character} & BAAD16 & 20 & 32 & 1e-2 \\ \cline{3-6}
                      &                            & BAAD6 & 10 & 128 & 1e-1,1e-2,1e-3 \\ \cline{1-6}
                      
% \hline
\end{tabular}
\label{fithyp}
\end{table}

\subsubsection{Classification}
The final downstream task is classification. The model is modified, as mentioned in Section \ref{classifier}. The encoder weights from the fine-tuned language model are loaded into the classifier. The decoder weights are randomly initialized. The dropout multiplier is set to 0.5, and the chosen learning rate is 1e-2 using a cyclical learning rate finder. Momentum values of 0.8 and 0.7 are used for optimization \cite{howard2018universal}. The entire model is frozen except the decoder, which is available for training. After training with the training set of the authorship attribution dataset for two epochs, a layer group is unfrozen and trained using sliced learning rates for two epochs. Slicing mainly distributes learning rates among the layers so that the initial layers are updated slowly to maintain the pre-trained weights. In contrast, the later layers, which are the most task-specific layers, are updated swiftly to learn the task at hand. A general rule of slicing learning rate has been followed where the slice consists of initial learning rate(lr) and $\frac{lr}{2.6^{4}}$ \cite{howard2018universal}. The unfreezing step is carried out one more time and trained for two epochs before unfreezing the entire model and training for six epochs. The training was stopped when it started to overfit. We used a batch size of 32 for all the models at this stage.
        
\section{Experiments}
\label{experiments}
Previous state-of-the-art models were re-created to compare the proposed models against baselines, and results were drawn in terms of accuracy and F1 score. Neither of the previous models uses transfer learning from entire models. They are based on using pre-trained character and word embeddings to boost performance. Based on experiments from \cite{ourchar}, and \cite{bhai}, we select the best performing models for word embeddings (using fastText skip-gram) and character embedding and evaluate them on BAAD16 and BAAD6 datasets. Besides, we also experiment with a transformer-based multilingual model called mBERT and a similar model trained only on Bangla text called Bangla-BERT. All the models used for comparison are discussed in brief below. Table \ref{experiemnts_params_table} shows the various hyper-parameters used for the models described below.

\subsection{CNN-LSTM Word level classifier}
The CNN-LSTM model \cite{ourchar} comprises of a mix of convolutional and LSTM layers suitable for training the corpus on word-level tokenization. The architecture consists of an embedding layer that is initialized with pre-trained word-embeddings using skip-gram of fastText. Vocabulary size of 60,000 and embedding vector size of 300 are used. Two convolutional and max pool layer pairs follow the embedding layer. The outputs from the convolutional layers are fed into an LSTM layer of 100 neurons, whose outputs are in turn fed into a fully connected layer of 512 neurons. Dropout is used in the fully connected layer to prevent the model from overfitting. A final softmax layer outputs probabilities for the classification. All parameters for the layers are kept the same as the originally proposed model.
The model is trained for 15 epochs with a learning rate of 0.001, decay of 1e-4, and batch size of 128. We used Adam optimizer and categorical cross-entropy as the loss function.

\subsection{CNN Character level classifier}
This model aims to use character signals as embedding and passes them on to several convolutional layers to extract information and later perform classification. The original CNN model \cite{ourchar} has been recreated and trained for the entire dataset. The first layer consists of the embedding layer whose weights are set as the pre-trained weights \cite{ourchar} obtained from training the news dataset mentioned in Section \ref{corpus}. The vocabulary and embedding size is 181. The vocabulary is the same as the one used with the proposed character-level AWD-LSTM model except for the special tokens. The embedding layer is connected to 4 sequentially connected convolutional layers. Each convolutional layer is followed by a max pool layer of kernel size 3. A fully connected layer is stacked on top with the number of nodes set as 512. The number of filters and kernel sizes for the convolutional layers is kept the same as in the original model. The last layer is the output layer with softmax activation.
With a batch size of 128 and a maximum text sample length of 3000, the model has been trained on the dataset for 15 epochs. A dropout rate of 0.5 is used in the fully connected layer as regularization. The rest of the hyper-parameters are kept consistent with the CNN-LSTM model.
%The learning rate was set to be 0.001 and decay of 1e-4. Optimizer and loss function remained to Adam and categorical cross-entropy.

\subsection{Multilingual BERT}
In recent times transfer learning with transformer-based models has shown promise in many NLP tasks. These models are large and are pre-trained on huge amounts of text in an unsupervised manner. One of the most widely used models is BERT (Bidirectional Encoder Representations from Transformers) \cite{devlin2018bert}. Multilingual BERT (mBERT) has Bangla as part of its training corpus, among 100+ other languages. BERT's architecture is a multi-layer bidirectional Transformer encoder based on the original Transformer \cite{vaswani2017attention}. BERT's uniqueness is that it uses bi-directional context instead of only left-right context as in autoregressive language modeling tasks. BERT uses WordPiece embeddings \cite{wu2016google} where a word maybe be broken into multiple pieces, and each piece is prefixed with the symbols \#\# to indicate their origins. The first token of all sequences is a special token [CLS]. The final hidden state of this token is used as an aggregate sequence representation. BERT has three kinds of embeddings: word embedding, positional embedding, and sentence embedding. And the final embedding of the model is the sum of these three embeddings. We used the small version of BERT which has L = 12 attention blocks, A = 12 self-attention heads and the hidden size is H = 768 \cite{devlin2018bert}. 

For classification, we only concern ourselves with the output of the [CLS] token present at the beginning of every sentence. We add a single-layer classifier on top of the BERT encoder. Specifically, the final hidden vector $C\in \rm I\!R^H$ corresponding to the input token [CLS] is input to the classifier layer, and the number of outputs is the number of class labels; in our case, the number of authors in the dataset. The entire model is then fine-tuned end-to-end along with the new layer. For our task of authorship attribution in Bangla, we use the trained, multilingual BERT model called \emph{bert-base-multilingual-cased}. We use batch size of 6 and total sequence length of 512 as per the original setting. Unlike the original BERT, we use Slanted Triangular learning rates, Discriminative Learning rates, and gradual unfreezing, similar to our proposed approach. We select the learning rates at various stages using a learning rate finder. Thus we choose an initial learning rate of 0.0004 for the BAAD16 dataset and train the classifier for 40 epochs. For the BAAD6 dataset learning rate of 0.0008 was selected and was trained for 35 epochs. The number of epochs was selected based on the validation set errors. When the validation set errors start to increase or oscillate, the training is stopped.

\subsection{Bangla-BERT}
\label{bbert}
BERT based models have been used extensively in various tasks recently. To this end, a BERT variation was trained purely in Bangla by Sarkar \cite{bbertsagor}. In contrast to mBERT, this model incorporates more Bangla training data than mBERT. It was trained on the Bangla common crawl corpus from OSCAR\footnote{https://oscar-corpus.com/} as well as the Bangla Wikipedia dump dataset. This uses the same \emph{bert-base} architecture as in mBERT and consists of 110M parameters. We use the \emph{sagorsarker/bangla-bert-base} checkpoint from the Hugging Face Transformers library for this purpose. For this model also, we use a batch size of 6 and a maximum sequence length of 512 as per the original model's settings. To train for the BAAD16 dataset, we choose an initial learning rate of $1.32E-6$ using the learning rate finder. The model was fine-tuned for a total of 40 epochs using gradual unfreezing, and the learning rate was adjusted using the learning rate finder at each unfreezing step. The training was stopped when the validation set performance started degrading, and the loss started increasing. For BAAD6, the initial learning rate was $1.74E-3$, and it was trained for a total of 30 epochs following similar unfreezing and learning rate adjustment methods.

\begin{table}[h]
\centering
\caption{Hyper-parameters of other models used for comparison.}
\label{experiemnts_params_table}
\begin{tabular}{|c|c|c|c|c|c|c|}
\hline
\textbf{\begin{tabular}[c]{@{}c@{}}Pre-training\\ dataset\end{tabular}} & \textbf{Model} & \textbf{Tokenization} & \textbf{\begin{tabular}[c]{@{}c@{}}Fine-tuning\\ dataset\end{tabular}} & \textbf{Epoch} & \textbf{\begin{tabular}[c]{@{}c@{}}Batch\\ size\end{tabular}} & \textbf{\begin{tabular}[c]{@{}c@{}}Learning\\ rate\end{tabular}} \\ \hline
\multirow{2}{*}{\begin{tabular}[c]{@{}c@{}}News\\ (fastText)\end{tabular}} & \multirow{2}{*}{CNN-LSTM} & \multirow{2}{*}{Word} & BAAD6 & 15 & 128 & 1e-3 \\ \cline{4-7} 
 &  &  & BAAD16 & 15 & 128 & 1e-3 \\ \hline
\multirow{2}{*}{\begin{tabular}[c]{@{}c@{}}News\\ (Embedding\\ layer)\end{tabular}} & \multirow{2}{*}{CNN} & \multirow{2}{*}{Character} & BAAD6 & 15 & 128 & 1e-3 \\[0.25cm] \cline{4-7} 
 &  &  & BAAD16 & 15 & 128 & 1e-3 \\[0.25cm] \hline
\multirow{2}{*}{Wiki dump} & \multirow{2}{*}{mBERT} & \multirow{2}{*}{\begin{tabular}[c]{@{}c@{}}WordPiece\\ (Sub-word)\end{tabular}} & BAAD6 & 35 & 6 & 8e-4 \\ \cline{4-7} 
 &  &  & BAAD16 & 40 & 6 & 4e-4 \\ \hline
\multirow{2}{*}{\begin{tabular}[c]{@{}c@{}}Wiki dump,\\ Commoncrawl\end{tabular}} & \multirow{2}{*}{Bangla-BERT} & \multirow{2}{*}{\begin{tabular}[c]{@{}c@{}}WordPiece\\ (Sub-word)\end{tabular}} & BAAD6 & 30 & 6 & 1.74\text{e-}3 \\ \cline{4-7} 
 &  &  & BAAD16 & 40 & 6 & 1.32\text{e-}6 \\ \hline
\end{tabular}
\end{table}

\section{Results and Discussions}
\label{results}

\subsection{Train and Test splits}
After training, the models were tested with the held-out test set. BAAD6 has a separate test set with 12.5\% data \cite{bhai}. The same test set was used for the purposes of consistency. BAAD16, on the other hand, holds out 20\% of the dataset from each author as a test set. This dataset was structured in line with the one used in \cite{ourchar} to be able to compare and draw from its results directly. Since the number of samples is not in the millions, test sizes smaller than 20\% are not representative, and choosing test sizes greater than 20\% reduces training data significantly since the training set is used for both tuning the language model as well as to learn classification.

\subsection{Metrics}
All the models were measured in terms of accuracy and F1 score. Although BAAD6 is balanced, BAAD16 is not (refer to Section \ref{corpus}), for which a look at the F1-score is also necessary. Accuracy measures the percentage of samples correctly identified, whereas the F1 score is the harmonic sum of the precision and recall. F1-score thus gives us the ability to look into pieces correctly identified from each class in a comparable form. Table \ref{clsresult} shows the summary of the results obtained from various experimented models against all the variations of the proposed model. %Figure \ref{prediction} shows a few sample texts and their actual and predicted authors using the best performing model.

%% Recheck Bangla-BERT F1s

%%%%%%%%%%%%%%%%%%%%%%%%%%%%%% result table Classification
\begin{table}[h]
\caption{Results of Classification}
{\renewcommand{\arraystretch}{1.08}% for the vertical padding
\begin{tabular}{|l|l|l|l|l|l|l|}
\hline
\textbf{Dataset}        & \textbf{Model}          & \textbf{Tokenization}              & \textbf{\begin{tabular}[c]{@{}l@{}}Pre-traning\\ Dataset\end{tabular}} & \textbf{\begin{tabular}[c]{@{}l@{}}FiT\\ Perplexity\end{tabular}} & \textbf{Accuracy \%}    & \textbf{F1 Score}        \\ \hline
\multirow{10}{*}{BAAD16} & \multirow{6}{*}{ULMFiT} & \multirow{2}{*}{Word}              & News                                                                   & 74.67                                                             & 99.58          & 0.9855          \\ \cline{4-7} 
                        &                         &                                    & Wiki                                                                   & 60.91                                                             & 99.67          & 0.9967          \\ \cline{3-7} 
                        &                         & \multirow{2}{*}{\textbf{Sub-word}} & \textbf{News}                                                          & 62.45                                                             & \textbf{99.80} & \textbf{0.9980} \\ \cline{4-7} 
                        &                         &                                    & Wiki                                                                   & 57.85                                                             & 99.72          & 0.9972          \\ \cline{3-7} 
                        &                         & \multirow{2}{*}{Character}         & News                                                                   & 3.42                                                              & 98.55          & 0.9855          \\ \cline{4-7} 
                        &                         &                                    & Wiki                                                                   & 3.36                                                              & 98.58          & 0.9858          \\ \cline{2-7} 
                        & Char-CNN                & Character                          & News                                                                      & -                                                                 & 86.28          & 0.7981          \\ \cline{2-7} 
                        & CNN-LSTM                & Word                               & News                                                                      & -                                                                 & 93.82          & 0.8934          \\ \cline{2-7} 
                        & mBERT                   & Sub-word                           & Wiki                                                                   & -                                                                 & 94.79          & 0.9283          \\ \cline{2-7} 
                        & Bangla-BERT             & Sub-word                           & \begin{tabular}[c]{@{}l@{}}Wiki,\\Common-crawl\end{tabular}              & -                                                                 & 96.80          & 0.9533          \\ \hline
\hline
\multirow{10}{*}{BAAD6}  & \multirow{6}{*}{ULMFiT} & \multirow{2}{*}{Word}                               & News                                                                   & 203.11                                                            & 94.67          & 0.9469          \\ \cline{4-7} 
                        &                         &                                    & Wiki                                                                   & 149.58                                                            & 94.33          & 0.9437          \\ \cline{3-7} 
                        &                         & \multirow{2}{*}{\textbf{Sub-word}} & \textbf{News}                                                          & 233.04                                                            & \textbf{95.33} & \textbf{0.9536} \\ \cline{4-7} 
                        &                         &                                    & Wiki                                                                   & 260.90                                                            & 94.67          & 0.9476          \\ \cline{3-7} 
                        &                         & \multirow{2}{*}{Character}         & News                                                                   & 4.47                                                              & 83.67          & 0.8360          \\ \cline{4-7} 
                        &                         &                                    & Wiki                                                                   & 3.71                                                              & 90.00          & 0.9007          \\ \cline{2-7} 
                        & Char-CNN                & Character                          & News                                                                      & -                                                                 & 73.33          & 0.7147          \\ \cline{2-7} 
                        & CNN-LSTM                & Word                               & News                                                                      & -                                                                 & 66.33          & 0.6320          \\ \cline{2-7} 
                        & mBERT                   & Sub-word                           & Wiki                                                                  & -                                                                 & 93.33          & 0.9301          \\ \cline{2-7} 
                        & Bangla-BERT             & Sub-word                           & \begin{tabular}[c]{@{}l@{}}Wiki,\\Commoncrawl\end{tabular}              & -                                                                 & 90.33          & 0.9012          \\ \hline
\end{tabular}
}
\label{clsresult}
\end{table}

\subsection{Results}
From Table \ref{clsresult}, we see that the proposed methods of transfer learning using language modeling with AWD-LSTM architecture outperform the previous high-performing models in Bangla literature in the target datasets. Specifically, the sub-word tokenized models have consistently performed well. The mBERT model performs well considering that it is pre-trained on over 100 languages, of which only a tiny portion consists of Bangla texts. But counter-intuitively, a pure Bangla-BERT model does not seem to outperform the mBERT model. One hypothesis could be the size of the dataset we use to train the model. The model may be over-parameterized for the current Authorship Attribution datasets, and training more and more epochs leads to a memorizing issue, ultimately causing the validation and test sets to perform poorly. Nevertheless, we only train up to the point where the validation loss stops decreasing, thus getting the best possible model checkpoint from the available datasets. 

\subsection{K-fold validation}

To test the robustness of our proposed systems, we perform K-fold cross-validation on BAAD6 and BAAD16 datasets with the best performing model, i.e, the News dataset pre-trained sub-word level ULMFiT model. We perform a 5-fold cross-validation and average the accuracies and F1 scores across the folds. Table \ref{kfold_table} shows the results along with the margin of error at a 95\% confidence interval. The authorship-attribution fine-tuned model for each dataset was used, i.e, we start from the \emph{classification} level of training. They were trained with an early stopping callback of 2 epochs. For both datasets and all folds, the newly initialized layer was trained for 2 epochs, followed by 2 epochs with a layer group unfrozen, followed by another 2 epochs with another layer group unfrozen. Then the completely unfrozen model was trained until the loss stopped decreasing. For BAAD6, it was a total of 10 epochs; for BAAD16, the model trained for about 10-13 epochs in various folds. Besides K-fold validation, more minor variations of BAAD16 were also tested separately. The results of this analysis can be found in Section \ref{num_auth_effect}.

\begin{table}[h]
\centering
\caption{Results of 5-fold cross-validation for the subword ULMFiT models on BAAD6 and BAAD16 datasets. The margin of error at 95\% confidence interval was reported alongside the mean accuracy and F1-scores}
\label{kfold_table}
\begin{tabular}{|c|c|c|c|c|c|}
\hline
\textbf{\begin{tabular}[c]{@{}c@{}}Pre-training\\ Datset\end{tabular}} & \textbf{Tokenization} & \textbf{Model} & \textbf{Dataset} & \textbf{Accuracy \%} & \textbf{F1 Score} \\ \hline
\multirow{2}{*}{News} & \multirow{2}{*}{Sub-word} & \multirow{2}{*}{ULMFiT} & BAAD6 & \begin{tabular}[c]{@{}c@{}}95.0952 \\ ±0.339 (±0.36\%)\end{tabular} & \begin{tabular}[c]{@{}c@{}}0.9511 \\ ±0.00347 (±0.37\%)\end{tabular} \\ \cline{4-6} 
 &  &  & BAAD16 & \begin{tabular}[c]{@{}c@{}}99.2597 \\ ±0.436 (±0.44\%)\end{tabular} & \begin{tabular}[c]{@{}c@{}}0.9924 \\ ±0.00466 (±0.47\%)\end{tabular} \\ \hline
\end{tabular}
\end{table}

\subsection{Comparative analysis}
In addition, we compare our proposed model to most existing research in Bangla on authorship attribution in Table \ref{dataset_comparison}, demonstrating that our technique has outstanding performance when considering the size of the corpus, length of samples utilized, and the number of authors categorized. Most works use self-created datasets with very few authors, possibly bordering on topic classification. Some authors tend to write on specific topics, so classification becomes very easy and unlike real-life scenarios, therefore achieving high accuracy. The publicly available dataset BAAD6 is a slightly noisy dataset and was used to evaluate the proposed system and some of the previous approaches. Our approach shows improved accuracy in this dataset at 95.33\% from the previous 92.9\% using fastText word embedding \cite{bhai}.

\begin{table}[H]
\caption{Comparison of the proposed system with existing works in Bangla authorship attribution (ascending order of the number of authors in the dataset). The * mark indicates that the work was replicated in this paper and applied to the respective dataset.}
\label{dataset_comparison}

{\renewcommand{\arraystretch}{1.05}% for the vertical padding
\begin{tabular}{|l|l|l|l|l|l|}
\hline
\multicolumn{1}{|c|}{\textbf{Paper/Dataset}}                                                                & \multicolumn{1}{c|}{\textbf{\begin{tabular}[c]{@{}c@{}}No. of\\ author\end{tabular}}} & \multicolumn{1}{c|}{\textbf{\begin{tabular}[c]{@{}c@{}}Average\\ word /\\ author\end{tabular}}} & \multicolumn{1}{c|}{\textbf{\begin{tabular}[c]{@{}c@{}}Average\\ samples /\\ author\end{tabular}}} & \multicolumn{1}{c|}{\textbf{Method}}                                                                      & \multicolumn{1}{c|}{\textbf{\begin{tabular}[c]{@{}c@{}}Accuracy\\ (\%)\end{tabular}}} \\ \hline
\multirow{2}{*}{Das et al. \cite{das2011author}}                                                                   & \multirow{2}{*}{3}                                                                    & \multirow{2}{*}{325k}                                                                           & \multirow{2}{*}{12}                                                                                & \begin{tabular}[c]{@{}l@{}}Probabilistic classification\\ unigram + vocabulary richness\end{tabular}      & 91.67                                                                                 \\ \cline{5-6} 
                                                                                                            &                                                                                       &                                                                                                 &                                                                                                    & \begin{tabular}[c]{@{}l@{}}Probabilistic classification\\ bigram + vocabulary richness\end{tabular}       & 100                                                                                   \\ \hline
\begin{tabular}[c]{@{}l@{}}Chakraborty\\ et al. \cite{chakraborty2012authorship}\end{tabular}                      & 3                                                                                     & 4921k                                                                                           & 150                                                                                                & Extracted features with SVM                                                                               & 83.3                                                                                  \\ \hline
\begin{tabular}[c]{@{}l@{}}Anisuzzaman\\ et al. \cite{anisuzzaman2018authorship}\end{tabular}                      & 3                                                                                     & 35k                                                                                             & \multicolumn{1}{c|}{-}                                                                             & N-grams with Naive Bayes                                                                                  & 95                                                                                    \\ \hline
\multirow{2}{*}{Phani et al. \cite{phani2017supervised}}                                                           & \multirow{2}{*}{3}                                                                    & \multirow{2}{*}{104k}                                                                           & \multirow{2}{*}{1000}                                                                              & \begin{tabular}[c]{@{}l@{}}Tfidf of N-grams of both word \&\\ character, stop words with SVM\end{tabular} & 98.93                                                                                 \\ \cline{5-6} 
                                                                                                            &                                                                                       &                                                                                                 &                                                                                                    & \begin{tabular}[c]{@{}l@{}}Topic modeling as features\\ with Naive Bayes\end{tabular}                     & 100                                                                                   \\ \hline
Islam et al. \cite{islam2018authorship}                                                                            & 5                                                                                     & \multicolumn{1}{c|}{-}                                                                          & 394                                                                                                & \begin{tabular}[c]{@{}l@{}}N-grams and extracted features\\ with MLP\end{tabular}                         & 89                                                                                    \\ \hline
Hossain et al. \cite{hossain2017stylometric}                                                                       & 6                                                                                     & 185k                                                                                            & 100+                                                                                               & \begin{tabular}[c]{@{}l@{}}N-grams and extracted features\\ with voting classifier\end{tabular}           & 90.67                                                                                 \\ \hline
\multirow{3}{*}{\begin{tabular}[c]{@{}l@{}}Chowdhury\\ et al. \cite{bhai} (BAAD6)\end{tabular}}                    & \multirow{9}{*}{6}                                                                    & \multirow{9}{*}{384k}                                                                           & \multirow{9}{*}{350}                                                                               & Word embedding with MLP                                                                                   & 85.46                                                                                 \\ \cline{5-6} 
                                                                                                            &                                                                                       &                                                                                                 &                                                                                                    & Word embedding with LSTM                                                                                  & 89.6                                                                                  \\ \cline{5-6} 
                                                                                                            &                                                                                       &                                                                                                 &                                                                                                    & Word embedding with CNN                                                                                   & 92.9                                                                                  \\ \cline{1-1} \cline{5-6} 
\multirow{3}{*}{\begin{tabular}[c]{@{}l@{}}Chowdhury\\ et al. \cite{chowdhury2018authorship} (BAAD6)\end{tabular}} &                                                                                       &                                                                                                 &                                                                                                    & \begin{tabular}[c]{@{}l@{}}Word embedding with \\ Naive Bayes\end{tabular}                                & 61                                                                                    \\ \cline{5-6} 
                                                                                                            &                                                                                       &                                                                                                 &                                                                                                    & Word embedding with SVM                                                                                   & 84.4                                                                                  \\ \cline{5-6} 
                                                                                                            &                                                                                       &                                                                                                 &                                                                                                    & \begin{tabular}[c]{@{}l@{}}Word embedding with\\ heirarchical classifier\end{tabular}                     & 80.8                                                                                  \\ \cline{1-1} \cline{5-6} 
\begin{tabular}[c]{@{}l@{}}Khatun\\ et al. \cite{ourchar} (BAAD6)*\end{tabular}                                    &                                                                                       &                                                                                                 &                                                                                                    & Character Embedding with CNN                                                                              & 73.3                                                                                  \\ \cline{1-1} \cline{5-6} 
\begin{tabular}[c]{@{}l@{}}Khatun\\ et al. \cite{ourchar} (BAAD6)*\end{tabular}                                    &                                                                                       &                                                                                                 &                                                                                                    & \begin{tabular}[c]{@{}l@{}}Word Embedding with\\ CNN-LSTM\end{tabular}                                    & 66.3                                                                                  \\ \cline{1-1} \cline{5-6} 
\textbf{\begin{tabular}[c]{@{}l@{}}Proposed Model\\ (BAAD6)\end{tabular}}                                   &                                                                                       &                                                                                                 &                                                                                                    & \textbf{\begin{tabular}[c]{@{}l@{}}Transfer learning using\\ Language Model\end{tabular}}                 & \textbf{95.33}                                                                        \\ \hline
Islam et al. \cite{islam2017automatic}                                                                             & 10                                                                                    & \multicolumn{1}{c|}{-}                                                                          & 312                                                                                                & \begin{tabular}[c]{@{}l@{}}N-grams and POS tags\\ with Random forests\end{tabular}                        & 96                                                                                    \\ \hline
\begin{tabular}[c]{@{}l@{}}Khatun\\ et al. \cite{ourchar} (BAAD16)*\end{tabular}                                   & \multirow{3}{*}{16}                                                                   & \multirow{3}{*}{842k}                                                                           & \multirow{3}{*}{1122}                                                                              & Character Embedding with CNN                                                                              & 86.28                                                                                 \\ \cline{1-1} \cline{5-6} 
\begin{tabular}[c]{@{}l@{}}Khatun\\ et al. \cite{ourchar} (BAAD16)*\end{tabular}                                   &                                                                                       &                                                                                                 &                                                                                                    & \begin{tabular}[c]{@{}l@{}}Word Embedding with\\ CNN-LSTM\end{tabular}                                    & 93.82                                                                                 \\ \cline{1-1} \cline{5-6} 
\textbf{\begin{tabular}[c]{@{}l@{}}Proposed Model\\ (BAAD16)\end{tabular}}                                  &                                                                                       &                                                                                                 &                                                                                                    & \textbf{\begin{tabular}[c]{@{}l@{}}Transfer learning using\\ Language Model\end{tabular}}                 & \textbf{99.8}                                                                         \\ \hline
\end{tabular}
}
\end{table}

Furthermore, our curated dataset, BAAD16, was also used to compare the performance of some of the previously published models by replicating those architectures (details in Section \ref{experiments}). Performance improves from appx. 94\% (of mBERT, CNN-LSTM, etc) to 99.8\%, which is a significant improvement.

\subsection{Error analysis}
A small number of false classifications were briefly analyzed to reveal that the model confuses similar writings of different authors. It is not uncommon for writers to follow another famous writer or any of their peers. Besides, some writers of the same period also tend to follow similar writing patterns. Despite these, most cases were in-fact random errors whose correlations were not found in this brief analysis, at least not manually. An in-depth analysis of the errors can help reveal more about what the models learn in terms of understanding the structure and semantics of a language which remains a scope for future study.

\vspace{6pt}

The effects of various other factors on the obtained results, such as architecture, tokenization, number of authors, sample distribution, and dataset used for pre-training, are analysed to test the robustness and scalability of the proposed system. These are briefly discussed below. 

\subsection{Model Effect}
To assess the effects of ULMFiT model for the task of authorship attribution, we train some previous state-of-the-art models, namely Char-CNN and word-level CNN-LSTM. From Table \ref{clsresult}, it is evident that the ULMFiT model outperforms the other models by a significant amount on both BAAD6 and BAAD16 datasets. This clearly shows that the transfer learning approach is effectively applicable to the task of authorship attribution. The model learns Bangla from language model training and tunes to authorial writing styles on fine-tuning. Thus, with the additional steps of teaching the model a language, it can better detect the text's original author. Besides this, the AWD-LSTM architecture, along with the training techniques employed, offers a strong base for the language model as well as the classifier making the transfer learning thereof highly effective \cite{howard2018universal}.

\subsection{Tokenization Effect}
Tokenization is an integral part of any NLP task. In Bangla, the most common way so far has been to tokenize by words. In this paper, three types of tokenizations are attempted, and their effects are analyzed on the task of author classification. As described in Section \ref{tokenization}, we have performed word, sub-word, and character level tokenization. In general, the character level models perform worse than both word and sub-word models. This reflects that the character-level model faces difficulty choosing the correct author because of the long stream of tokens it has to go through before reaching any conclusion. LSTM layers pass one character at a time, making any sentence a long set of tokens. Therefore, the LSTM layers may have trouble gathering enough data about each author's difference in text or style. 

\begin{figure}[h]
\begin{subfigure}{.495\textwidth}
  \centering
  % include first image
  \includegraphics[width=1\linewidth]{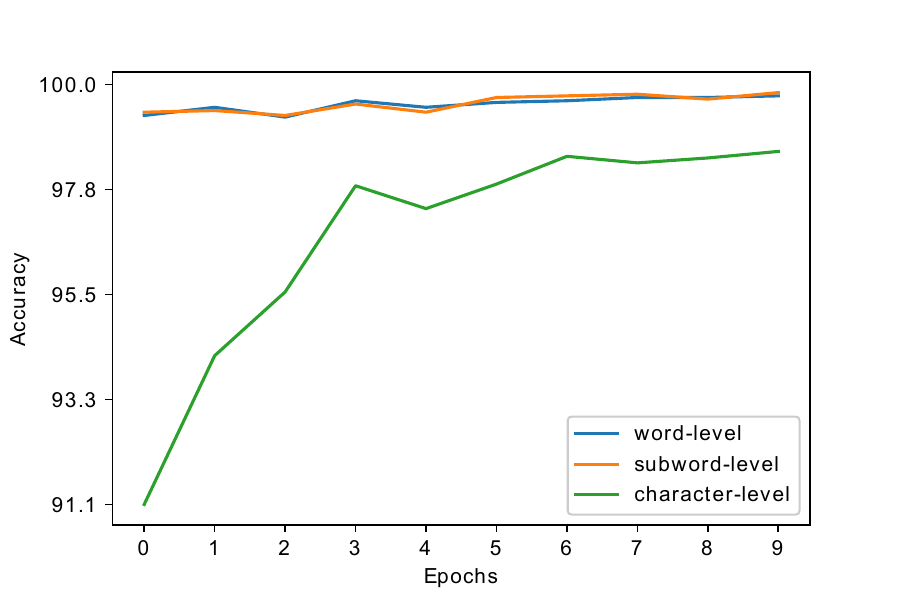}  
  \caption{On BAAD16}
%   \label{fig:sub-first}
\end{subfigure}
\begin{subfigure}{.495\textwidth}
  \centering
  % include second image
  \includegraphics[width=1\linewidth]{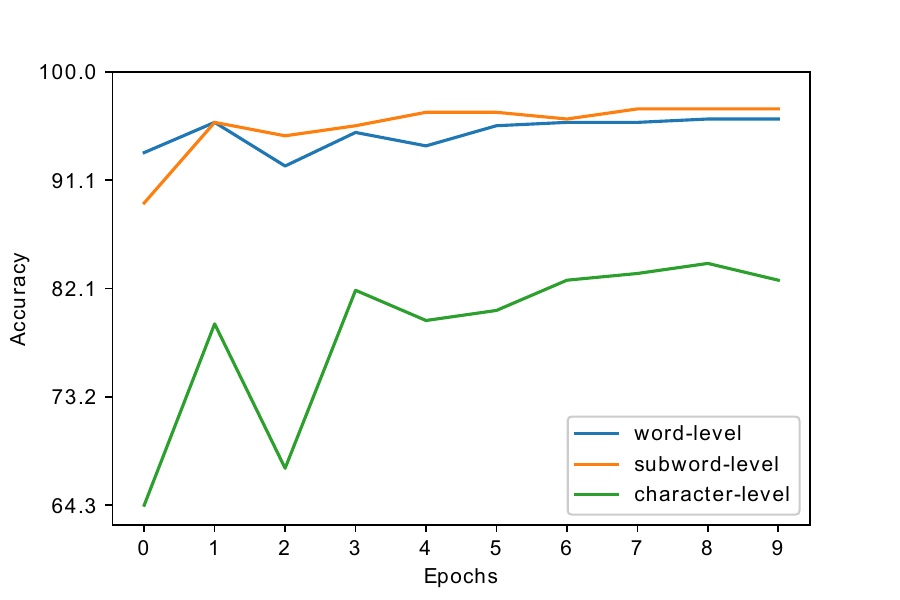}  
  \caption{On BAAD6}
%   \label{fig:sub-second}
\end{subfigure}
\caption{The accuracy versus epoch plot shows the models' performance comparison with different tokenizations. Here the character tokenized model cannot reach the level of subword and word tokenized models in both the tested datasets (BAAD6 and BAAD16). Notice the vertical axis does not start from zero for clarity.}
\label{tokenizationeffect}
% \end{figure}

% \begin{figure}[ht]
\begin{subfigure}{.495\textwidth}
  \centering
  % include first image
  \includegraphics[width=1\linewidth]{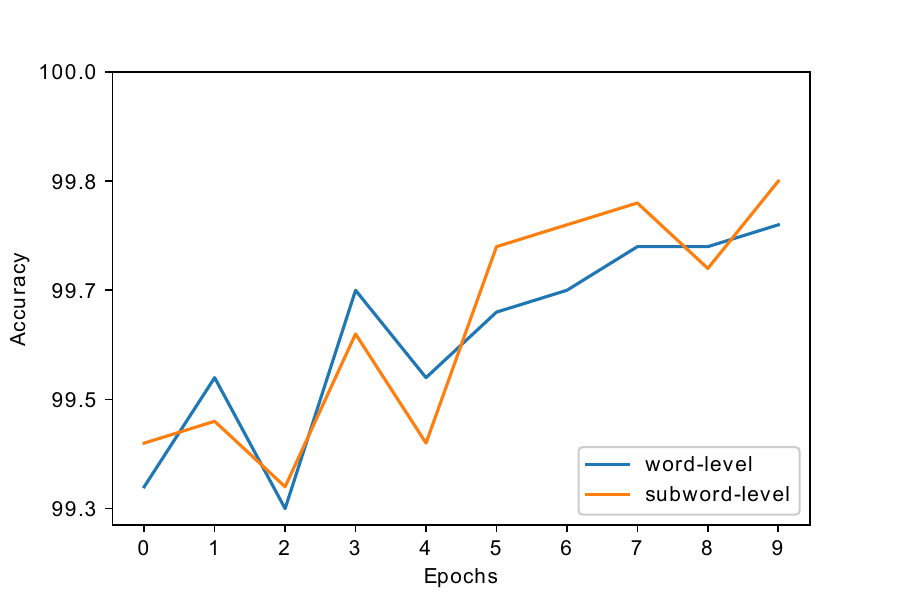}  
  \caption{On BAAD16}
%   \label{fig:sub-first}
\end{subfigure}
\begin{subfigure}{.495\textwidth}
  \centering
  % include second image
  \includegraphics[width=1\linewidth]{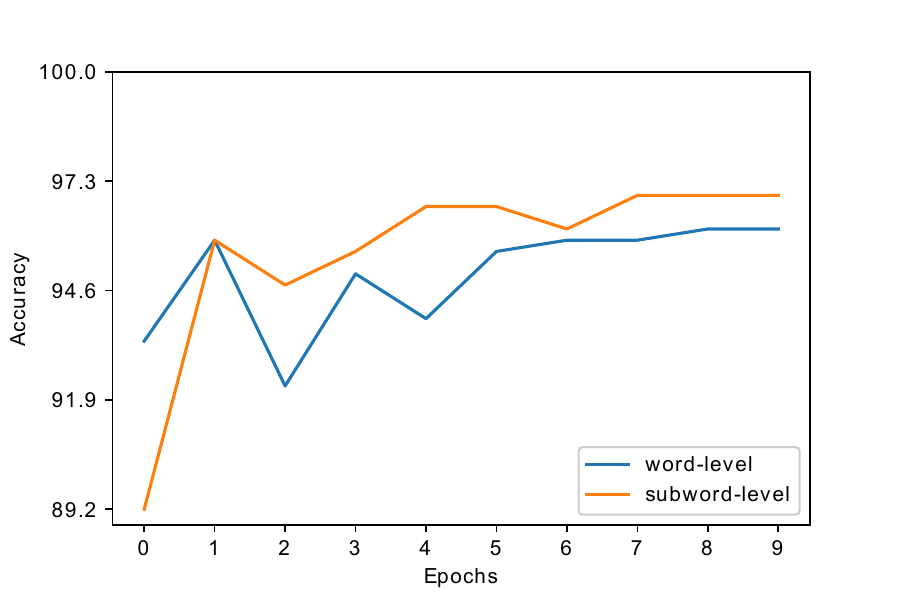}  
  \caption{On BAAD6}
%   \label{fig:sub-second}
\end{subfigure}
\caption{The accuracy versus epoch plot shows the models' performance comparison with word and subword tokenizations. It is an extended version of the previous figure, showing the two models advancing exceptionally closely together. For clarity, the vertical axis does not begin at zero.}
\label{word,subword tokenization}
\end{figure}

The word and sub-word level models perform nearly equally well, but the sub-word models begin to outrun the word-level models on further epochs. This information can be drawn from Figure \ref{tokenizationeffect}, where we see the progress for each tokenization type's model on both BAAD16 and BAAD6 in a zoomed-in scale for easy understanding. A more zoomed-in graph in Figure \ref{word,subword tokenization} shows the rising accuracy of the sub-word level model on each epoch, indicating a specific comparison of word and sub-word level models. The sub-word tokenization breaks down the text into multiple parts but not wholly. As a result, linguistic information about the words is kept intact, and it also provides information about the relationship between structural components of words. For example, it separates inflected prefixes or suffixes from words, so the model retains root word information as well as gets additional information from the separated parts. Whereas when tokenized by words, the relationship between related words with slight structure variations is not easily recognized.

\subsection{Effect of number of Authors}
\label{num_auth_effect}
\begin{figure}[H]
\centering
\includegraphics[width=0.8\linewidth]{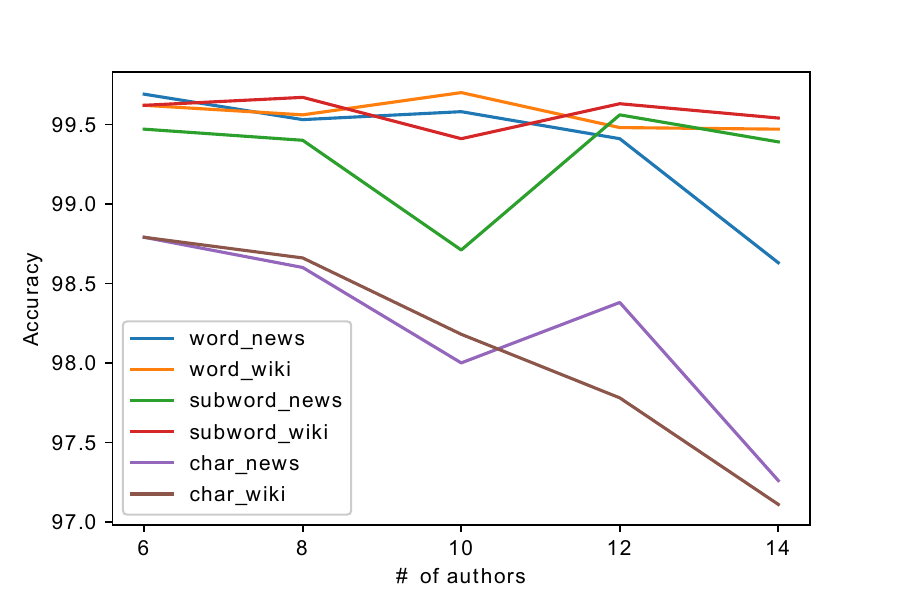}  
\caption{Performance of different models with increasing number of authors. The models vary by tokenization and the dataset it was pre-trained on, depicted in the legend in the form 'tokenization\_pretraining-dataset'. The vertical axis shows the classification accuracy, and the horizontal axis indicates the number of authors taken from the BAAD16 dataset.}
\label{authfig}
\end{figure}

In this section, the proposed models' results are compared with varying numbers of authors to determine the models' effectiveness against increasing classes trained with fewer samples. BAAD16 contains 16 authors in an imbalanced manner. It is chunked into five parts containing 6, 8, 10, 12, and 14 authors, respectively, randomly taking subsets from the original dataset. The samples per class in the derived datasets are truncated to the minimum number of samples among the classes. We have only compared the accuracy because all the models are now being trained and tested on balanced datasets. Accuracy is measured on 20\% held out dataset for each case, after training on 80\% of the data. Each of the six pre-trained models of 3 types (word, sub-word, and character level) is fine-tuned on these five sub-datasets. As we increase the number of authors, the sample per author also decreases, making it difficult for any deep learning model to learn each class of text. The summary is presented in Table \ref{authresult}. A graph depicting the accuracy trend with increasing number of authors for all the models is shown in Figure \ref{authfig}. 
% \clearpage

%%%%%%%%%%%%%%%%%%%%%%%%%%%%%% Author number variation result Classification
\begin{table}[h]
\caption{Accuracy of authorship attribution for increasing number of Authors. All subsets of the dataset are balanced, and the samples per author are displayed underneath the author number in each column.}
\begin{tabular}{|c|c|c|c|c|c|c|}
\hline
\multicolumn{2}{|l|}{}                                                                                                            & \multicolumn{5}{c|}{\textbf{\begin{tabular}[c]{@{}c@{}}\# of authors on the subset\\ \# of samples per author\end{tabular}}} \\ \hline
\multirow{2}{*}{\textbf{Tokenization}} & \multirow{2}{*}{\textbf{\begin{tabular}[c]{@{}c@{}}Pre-training\\ Dataset\end{tabular}}} & 6 authors               & 8 authors               & 10 authors             & 12 authors             & 14 authors             \\
                                       &                                                                                          & 1100                    & 931                     & 849                    & 562                    & 469                    \\ \hline
\multirow{2}{*}{Word}                  & News                                                                                     & \textbf{99.69}          & 99.53                   & 99.58                  & 99.41                  & 98.63                  \\ \cline{2-7} 
                                       & Wiki                                                                                     & 99.62                   & 99.56                   & \textbf{99.70}         & 99.48                  & 99.47                  \\ \hline
\multirow{2}{*}{Sub-word}              & News                                                                                     & 99.47                   & 99.40                   & 98.71                  & 99.56                  & 99.39                  \\ \cline{2-7} 
                                       & Wiki                                                                                     & 99.62                   & \textbf{99.67}          & 99.41                  & \textbf{99.63}         & \textbf{99.54}         \\ \hline
\multirow{2}{*}{Character}             & News                                                                                     & 98.79                   & 98.60                   & 98.00                  & 98.38                  & 97.26                  \\ \cline{2-7} 
                                       & Wiki                                                                                     & 98.79                   & 98.66                   & 98.18                  & 97.78                  & 97.11                  \\ \hline
\end{tabular}
\label{authresult}
\end{table}

Out of the five subsets created, the sub-word model pre-trained on the Wikipedia corpus performs best in the majority of cases (3 in this occasion). More importantly, as the sample number decreases, all other models more or less start performing worse than the sub-word tokenized model trained on the Wikipedia corpus. Figure \ref{authfig} shows the decline in the character model trained with lesser samples. Moreover, the word and sub-word level wiki models consistently perform better, and the sub-word model tends to give higher accuracy. From these results, we can conclude that the Wikipedia pre-trained sub-word model shows more stability than others. The reason behind this could be two-fold. Firstly the Wikipedia dataset has more varied text than the news dataset. This helps the model generalize better. Secondly, the sub-word level model provides enough information about word and sentence structure yet does not break it down too much (into characters) to lose sight of the overall picture of the sampled text.

\subsection{Effect of pre-training datasets}
\begin{figure}[H]
\begin{subfigure}{.495\textwidth}
  \centering
  % include first image
  \includegraphics[width=1\linewidth]{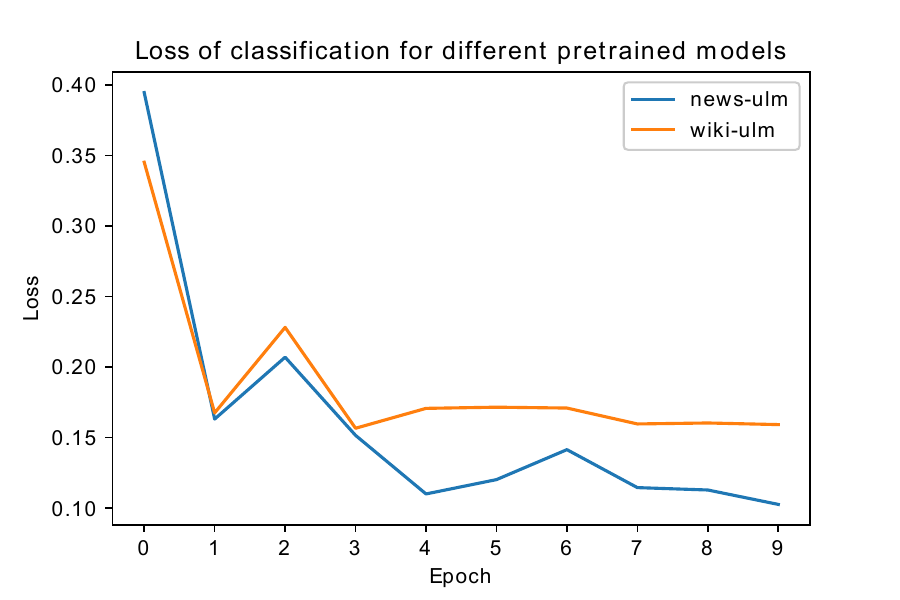}  
  \caption{On BAAD16}
%   \label{fig:sub-first}
\end{subfigure}
\begin{subfigure}{.495\textwidth}
  \centering
  % include second image
  \includegraphics[width=1\linewidth]{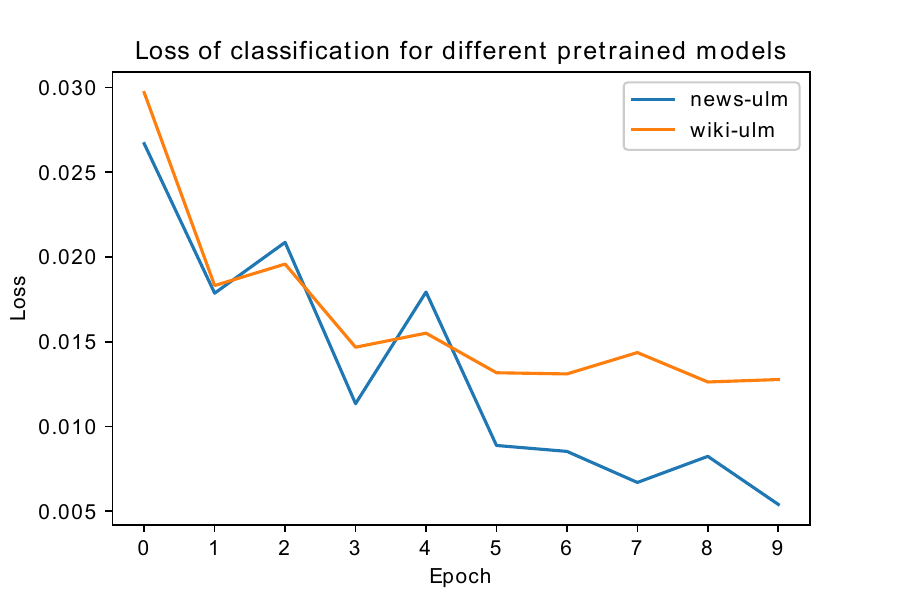}  
  \caption{On BAAD6}
%   \label{fig:sub-second}
\end{subfigure}
\caption{Loss versus epoch plot showing the effect of the dataset used in pre-training on classification. The news pre-trained model achieves lower loss for the task of authorship attribution in our case. Both the models are sub-word tokenized since sub-word tokenization achieves better performance than others.}
\label{wikivsnews}
\end{figure}

Although trained models with the Wikipedia dataset show better downstream generalization, the scenario changes in real-world applications. Data is often imbalanced, with some authors having very few training samples compared to others. The classifier's loss for each epoch is plotted to compare the effects of news and Wikipedia pre-trained subword models for each target dataset in Figure \ref{wikivsnews}. 

We see that the trained model with the news dataset works significantly better in terms of classifier's loss for both BAAD6 and BAAD16 datasets. The loss tends to decrease consistently with each epoch in both situations, although the news model's loss is substantially smaller than the wiki model's. It shows that the news dataset's categories and size enabled the classifier to distinguish better among the authors.

\section{Released Pre-trained Language Models}
\label{pretrained}

%%new wrote
A set of language models have been pre-trained on two different datasets, namely news and Wikipedia, as part of our workflow. Language models' performance is measured in terms of perplexity. This measure captures the degree of uncertainty in predicting the next word of the sentence. It is calculated as the exponentiation of the obtained loss. Low perplexity is a sign of a well-trained model. Table \ref{lmresult} lists the perplexities of the pre-trained models. We see that the character-level models perform significantly better for the task of text generation. An illustration of the produced text can be seen in Figure \ref{textfig}. The next best performance is obtained by the sub-word level model being slightly lower than the word-level models. The reason could be that the tokens' representation in smaller forms (e.g. characters) may provide additional information about the words and relationships among the words than the larger chunks can provide. Nevertheless, such information is not reliable for downstream tasks, as evident from our experiments where the character-level models underperformed compared to their counterparts.

All the pre-trained models have been made available for use and can be applied to any other Bangla NLP task of choice. The heavy task of pre-training has already been done. The weights have to be loaded, and the final dense layer has to be changed to perform the downstream task to use these models. It is to be noted that the downstream task has to be done in steps and preferably following the discriminative learning rate procedure so that the pre-trained weights can be efficiently utilized.
%%%%%%%%%%%%%%%%%%%%%%%%%%%%%%%%%% result table ULM
\begin{table}[h]
\centering
\caption{Perplexity of the trained Language Models.}
\begin{tabular}{| l | l | l | l |}
\hline
\textbf{Dataset} & \textbf{Tokenization} & \textbf{Perplexity}\\
\hline

\multirow{3}{*}{News} & Word        & 47.52 \\ \cline{2-3}
                      & Sub-word    & 43.61 \\ \cline{2-3}
                      & Character   & 3.20 \\ \cline{1-3}
\multirow{3}{*}{Wiki} & Word        & 83.62 \\ \cline{2-3}
                      & Sub-word    & 74.20 \\ \cline{2-3}
                      & Character   & 3.08 \\ \cline{1-3}

% \hline
\end{tabular}
\label{lmresult}
\end{table}

\section{Conclusion and future work}
\label{conslusion}
Transfer learning has been proven beneficial for domains with inadequate data labelling as it is first pre-trained on a resource-rich source domain before being fine-tuned on a downstream task. Therefore, in this work, we analyze transfer learning applications on authorship attribution using language modeling in three phases: Pre-training, Fine-tuning, and Author prediction. In this article, we present an AWD-LSTM and transfer learning-based scalable framework that addresses the problem of scarce language resources, scalability, and manual feature engineering for AABL. Towards that goal, experiments were conducted using the word, sub-word, and character level tokenization, demonstrating the efficacy of our approach with different tokenization. Furthermore, comparative analysis shows that our approach outperforms all existing traditional methods. To test our model with a larger dataset and longer text, we build a dataset containing the literary works of 16 authors with more than 13.4 million words in total. Our contribution to this work includes the proposal of a robust and scalable system for AABL using transfer learning, a comprehensive analysis of the effectiveness of different tokenization, a standard dataset, and the release of the pre-trained models for further use. The proposed method provides stability with fewer data and a large pool of authors. We conclude that the sub-word tokenized models consistently perform better regardless of the number of samples from our experiments. We want to investigate the application of transfer learning on deep CNN-based architectures and compare them with LSTM architecture with an attention model \cite{al2021identifying} for further work due to its pattern detection ability and faster training time. We would also like to make use of various transformer-based models for this task. Additionally,  cross-lingual authorship attributions and improvement of existing models on such tasks also interest us.

\bibliographystyle{ACM-Reference-Format}
\bibliography{main}

%%
%% If your work has an appendix, this is the place to put it.
% \appendix
% \section{Research Methods}
% \subsection{Part One}
% \subsection{Part Two}
% \section{Online Resources}

\end{document}